\documentclass[journal,twocolumn]{IEEEtran}

\usepackage{times}
\usepackage{epsfig}
\usepackage{graphicx}
\usepackage{amsmath}
\usepackage{amssymb}
\usepackage{algorithmic}
\usepackage[linesnumbered,ruled,vlined]{algorithm2e}
\usepackage{multirow}
\usepackage{booktabs}
\usepackage{siunitx}
\usepackage{bm}
\usepackage{amsfonts}
\usepackage{caption}

\usepackage{CJK}
\newif\ifdraft\drafttrue
\ifdraft

\newcommand\liming[1]{{\huge \color{black}#1  \textbf{}}}
\fi
\usepackage[pagebackref=true,breaklinks=true,letterpaper=true,colorlinks,bookmarks=false]{hyperref}

\ifCLASSINFOpdf

\else

\fi

\begin{document}
\begin{CJK*}{UTF8}{gkai}

\title{ \liming{Attention Regularized Laplace Graph for Domain Adaptation}}

\author{Lingkun Luo, ~Liming~Chen,~\IEEEmembership{Senior~member,~IEEE,  Shiqiang Hu}      

	\thanks{K. Luo and S. Qiang are from the School of Aeronautics and Astronautics, Shanghai Jiao Tong University, 800 Dongchuan Road, Shanghai, China e-mail: (lolinkun1988,sqhu)@sjtu.edu.cn}
\thanks{L. Chen from LIRIS, CNRS UMR 5205, 	Ecole Centrale de Lyon, 36 avenue Guy de Collongue, Ecully,  France e-mail: liming.chen@ec-lyon.fr.}
\thanks{This work for Liming Chen was in part supported by the 4D Vision project funded by the Partner University Fund (PUF), a FACE program, as well as the French Research Agency, l’Agence Nationale de Recherche (ANR), through the projects Learn Real ( ANR-18-CHR3-0002-01 ), Chiron (ANR-20-IADJ-0001-01), Aristotle (ANR-21-FAI1-0009-01), and the joint support of the French national program of investment of the future and the regions through the PSPC FAIR Waste project. This work was supported by the National Natural Science Foundation of China (61773262，62006152), and the China Aviation Science Foundation (20142057006). }
\thanks{This work is accepted by IEEE Transactions on Image Processing and will be available online soon.}
}

\markboth{Journal of \LaTeX\ 2021}%
{Shell \MakeLowercase{\textit{et al.}}: Bare Demo of IEEEtran.cls for IEEE Journals}

\maketitle

\begin{abstract}
In leveraging manifold learning in domain adaptation \textbf{(DA)}, graph embedding-based \textbf{DA} methods have shown their effectiveness in preserving data manifold through the Laplace graph. However, current graph embedding \textbf{DA} methods suffer from two issues: 1). they are only concerned with preservation of the underlying data structures in the embedding and ignore sub-domain adaptation, which requires taking into account intra-class similarity and inter-class dissimilarity, thereby leading to negative transfer; 2). manifold learning is proposed across different feature/label spaces separately, thereby hindering unified comprehensive manifold learning. In this paper, starting from our previous \textbf{DGA-DA}, we propose a novel \textbf{DA} method, namely \textit{A}ttention \textit{R}egularized Laplace \textit{G}raph-based \textit{D}omain \textit{A}daptation (\textbf{ARG-DA}), to remedy the aforementioned issues. Specifically, by weighting the importance across different sub-domain adaptation tasks, we propose the \textit{A}ttention \textit{R}egularized Laplace \textbf{G}raph for class aware \textbf{DA}, thereby generating the attention regularized \textbf{DA}. 	Furthermore, using a specifically designed \textbf{FEEL} strategy, our approach dynamically unifies alignment of the manifold structures across different feature/label spaces, thus leading to comprehensive manifold learning. Comprehensive experiments are carried out to verify the effectiveness of the proposed \textbf{DA} method, which consistently outperforms the state of the art \textbf{DA} methods on 7 standard \textbf{DA} benchmarks, \textit{i.e.}, 37 cross-domain image classification tasks including object, face, and digit images.  An in-depth analysis of the proposed \textbf{DA} method is also discussed, including sensitivity, convergence, and robustness.
\end{abstract}

%
\IEEEpeerreviewmaketitle

\vspace{-5mm}

	\section{Introduction}
	\label{Introduction}

\IEEEPARstart{S}{upervised} learning was at the forefront of increasing real-world applications within the big data era  \cite{pan2010survey, 7078994,DBLP:journals/tnn/ShaoZL15,DBLP:journals/csur/LuLHWC20}. However, its success relied greatly on the availability of large amounts of high-quality labeled data and assumed that training and testing data share the same distribution for the effectiveness of the learned model. Unfortunately, in real-life applications, data distribution shifts, due to factors as diverse as environment variations, \textit{e.g.}, lighting, temperature, differences in data capture devices, \textit{etc.}, frequently (though not always) occur. It is thus of capital importance to guarantee that a learned model using a given set of training data, namely the \textit{source domain}, remains effective on a set of testing data, namely the \textit{target domain}, despite data distribution shifts. This is precisely the objective of the research field of domain adaptation (\textbf{DA}), which aims to develop theories and techniques for effective learning algorithms, while making use of previously labeled data (source domain) and unlabeled data in the current task (target domain), despite the distribution divergence.

While \textbf{DA} can be \textit{semi-supervised} by assuming a certain amount of labeled data are available in the target domain, in this paper we are interested in \textit{unsupervised} \textbf{DA} where we assume that the target domain has no labels. While there also exists an increasing number of deep learning-based unsupervised DA methods, here we focus on \textit{shallow} DA methods as they are easier to train and can provide insights into the design decisions of deep DA methods. The relationships between \textit{shallow} and \textit{deep} DA methods will be further discussed in depth in Sect.\ref{Related Work} on related works. 

State-of-the-art \textit{shallow} DA methods can be categorized into \textit{instance}-based \cite{pan2010survey,donahue2013semi}, \textit{feature}-based  \cite{Busto_2017_ICCV,long2013transfer,DBLP:journals/tip/XuFWLZ16}, or \textit{classifier}-based. Classifier-based DA is widely applied in semi-supervised DA as it aims to fit a classifier trained on the source domain data to target domain data through adaptation of its parameters, thereby requiring some labels in the target domain \cite{tang2017visual}. The instance-based approach generally assumes that 1) the conditional distributions of the source and target domains  are identical \cite{Zhang_2017_CVPR}, and 2) a certain portion of the data in the source domain can be reused \cite{pan2010survey} for learning in the target domain through re-weighting. Feature-based adaptation \cite{sun2016return,DBLP:conf/cvpr/HerathHP17,ganin2016domain,tzeng2017adversarial, luo2020discriminative} relaxes such a strict assumption and only requires that there exists a mapping from the input data space to a latent shared feature representation space. This latent shared feature space captures the information necessary for training classifiers for both source and target tasks. In this paper, we propose a novel \textit{feature-based} \textbf{DA} method, which searches a latent feature space using the Laplace graph, but which is regularized by class aware attention along with a unified manifold alignment framework for an effective \textbf{DA}.

A common method for approaching feature adaptation is to seek a shared latent subspace between the source and target domains via optimization \cite{7078994,Busto_2017_ICCV}. State-of-the-art features three main lines of approach, namely, data geometric structure alignment-based (\textbf{DGSA}), data distribution centered (\textbf{DDC}) or their hybridization. \textbf{DDC} methods  \cite{luo2020discriminative, DBLP:conf/icml/LongZ0J17, liang2018aggregating,lu2018embarrassingly} aim to search a latent subspace where the discrepancy between the source and target data distributions is minimized, via various distances, \textit{e.g.}, Bregman divergence \cite{si2010bregman}, Geodesic distance \cite{gong2012geodesic}, Wasserstein distance \cite{courty2017joint, courty2017optimal} or Maximum Mean Discrepancy \cite{gretton2007kernel} (MMD). \textbf{DGSA}-based methods seek to align the underlying data geometric structures and can be further categorized into  subspace alignment-based \textbf{DA} (\textbf{DGSA-SA DA}) and graph embedding-based \textbf{DA} (\textbf{DGSA-GE DA}). \textbf{DGSA-SA} methods, \textit{e.g.},  \cite{sun2016return,DBLP:journals/ijcv/ShaoKF14,DBLP:journals/tip/XuFWLZ16}, aim to align different feature spaces under the reconstruction framework using the low rank constraint and/or sparse representation, while \textbf{DGSA-GE} methods explicitly model the underlying data geometric structure by using Laplace graph representation-based manifold learning, thereby seeking a subspace where the source and target data can be well aligned and interleaved by preserving inherent hidden geometric data structures. By hybridizing the \textbf{DGSA-GE} and \textbf{DDC} approaches, our proposed method, namely \textbf{ARG-DA}, strives to leverage the merits of simultaneous alignment of both data distributions and the underlying data geometric structures between the source and target domains.

\begin{figure}[h!]

	\centering
	\includegraphics[width=1\linewidth]{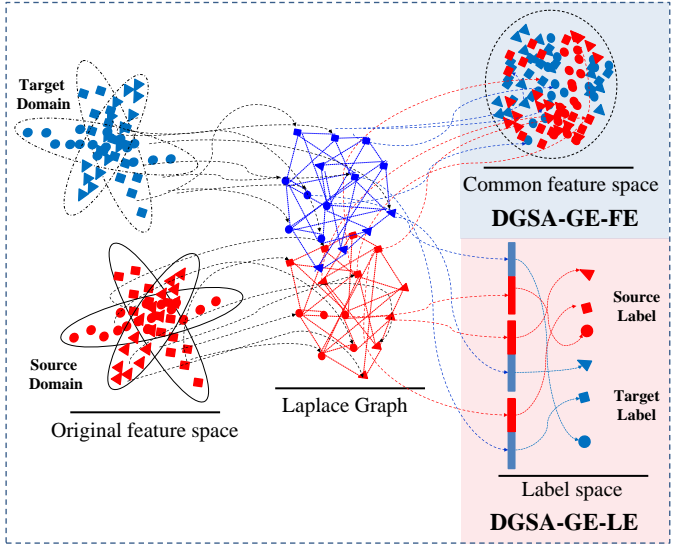}
	\caption{ Illustration of difference across the \textbf{F}eature space \textbf{E}mbedding-based \textbf{DA} \textbf{(DGSA-GE-FE)} and the \textbf{L}abel space \textbf{E}mbedding-based \textbf{DA} \textbf{(DGSA-GE-LE)}.} 
	\label{fig:1}
\end{figure}

Popular \textbf{DGSA-GE} methods align different feature/label spaces using graph embedding techniques and can be further summarized into two main research lines, namely \textit{feature space} embedding-based \textbf{DA} (\textbf{DGSA-GE-FE}),  \textit{e.g.}, \cite{jing2020adaptive,zhang2019manifold}, and \textit{label space} embedding-based \textbf{DA} (\textbf{DGSA-GE-LE}), \textit{e.g.}, \cite{zhang2016robust,li2020discriminative,luo2020discriminative}. As shown in Fig.\ref{fig:1}, \textbf{DGSA-GE-FE} methods seek a common feature space between the source and the target domain by preserving the underlying data geometric structure, while \textbf{DGSA-GE-LE} methods enforce the alignment of manifold structures across the feature space and the label space.

\begin{figure}[h!]
	\centering
	\includegraphics[width=1\linewidth]{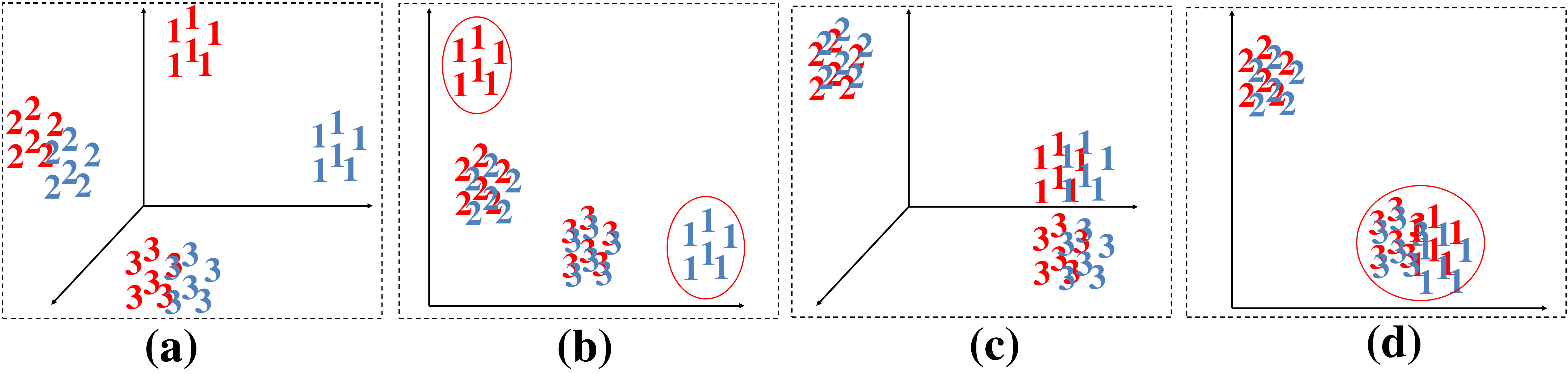}
	
	\caption {Fig.\ref{fig:4}.(a,c) represents the feature representation in its original three-dimensional feature space; Fig.\ref{fig:4}.(b,d) represents the feature representation in its two-dimensional common feature space by \textbf{DGSA-GE}; red and blue samples represent the source and target domains, while the numbers '1', '2', and '3' denote three different classes or sub-domains, respectively.} 
	\label{fig:4}

\end{figure}

To gain insights into both \textbf{DGSA-GE-FE} and \textbf{DGSA-GE-LE} methods, we review the cornerstone theoretical result in \textbf{DA}  \cite{ben2010theory,kifer2004detecting}, which estimates an error bound of a learned hypothesis $h$ on a target domain as follows:  
\vspace{-4.75mm}

\begin{equation}\label{eq:bound}
	\resizebox{0.9\hsize}{!}{%
		$\begin{array}{l}
		{e_{\cal T}}(h) \le {e_{\cal S}}(h) + {d_{\cal H}}({{\cal D}_{\cal S}},{{\cal D}_{\cal T}})+ \\
		\;\;\;\;\;\;\;\; \min \left\{ {{{\cal E}_{{{\cal D}_{\cal S}}}}\left[ {\left| {{f_{\cal S}}({\bf{x}}) - {f_{\cal T}}({\bf{x}})} \right|} \right],{{\cal E}_{{{\cal D}_{\cal T}}}}\left[ {\left| {{f_{\cal S}}({\bf{x}}) - {f_{\cal T}}({\bf{x}})} \right|} \right]} \right\}
		\end{array}$}
\end{equation}
\vspace{-10pt} 

where ${e_{\cal S}}(h)$ denotes the classification error on the source domain, and ${{d_{\cal H}}({{\cal D}_{\cal S}},{{\cal D}_{\cal T}})}$ measures the $\mathcal{H}$\emph{-divergence} \cite{kifer2004detecting} between two distributions ($\mathcal{D_S}$, $\mathcal{D_T}$) of the source and target domain. The last term of Eq.(\ref{eq:bound}) represents the difference in labeling functions across the two domains.
Following Eq.(\ref{eq:bound}), both the \textbf{DGSA-GE-FE} and \textbf{DGSA-GE-LE} methods are concerned with manifold structure preservation, which potentially shrinks the divergence between different labeling functions proposed on the two domains, thereby reducing term.3 of Eq.(\ref{eq:bound}). Different from \textbf{DGSA-GE-FE}, \textbf{DGSA-GE-LE} methods further leverage the discriminative power induced by the labels available on the source domain data by aligning the underlying data geometric structure in accordance with the data labels, thereby reducing the classification error on the source domain ${e_{\cal S}}(h)$ (term.1). However, despite interesting performance enabled \textbf{DGSA-GE} methods, they still suffer from the following two issues:




\textbf{Lack of Attention}:  \textcolor{red}{The \textbf{attention} mechanism refers to pooling a sequence or a set of features with different weights in order to compute proper representation of the whole sequence or set   \cite{vaswani2017attention,hu2018squeeze}, thereby making it possible to deal differently with different data features while aggregating.} However, by enforcing the maintenance of the geometric structure of the underlying data manifold while searching a joint feature subspace between the source and target domains, Laplace graph-based embedding \textbf{DA} methods, \textit{i.e.}, \textbf{DGSA-GE} methods, \textcolor{red}{treat equally all the sub-domain adaptations} and can lead to cross-domain misalignment as shown in Fig.\ref{fig:4}. As can be seen, Fig.\ref{fig:4}.(b,d) depicts the two-dimensional embeddings from their original three-dimensional feature space as shown in Fig.\ref{fig:4}.(a,c), respectively.


\begin{itemize}
	\item \textbf{P1:}  Fig.\ref{fig:4}.(b) preserves the data geometric structure of the original data space as in Fig.\ref{fig:4}.(a) but as a result fails to close adaptatively the corresponding sub-domains, \textit{e.g.},  '1', since  the two sub-domains ('1') are  distant in the original data manifold as depicted in Fig.\ref{fig:4}.(a).

	\item \textbf{P2:} Fig.\ref{fig:4}.(d) succeeds in aligning the data manifold structures with respect to their original data space as in Fig.\ref{fig:4}.(c), but as a result fails to separate the two sub-domains, namely, '1' and '3', which are geometrically close in the original data space.

	
\end{itemize}

The failures of \textbf{P1} and \textbf{P2} are due to the fact that traditional graph embedding-based manifold learning only seeks to \textbf{preserve} the relative distances across different feature spaces.  However, solving a  \textbf{DA} task requires bringing closer intra sub-domains across domains, while increasing the relative distances of inter sub-domains in the novel projected feature space. As a result, in this paper we propose a novel discriminativeness aware graph embedding, namely, Attention Regularized Graph (\textbf{ARG}), to remedy the aforementioned issues as highlighted by \textbf{P1} and \textbf{P2}.

\textbf{Incomplete Unification}: by aligning the manifold structures across different feature spaces, \textbf{DGSA-GE-FE} methods harness the geometric knowledge transferred across feature spaces, but fail to leverage the discriminative power induced by the labels available in the source domain data. On the other hand, \textbf{DGSA-GE-LE} remedies the previous issue and unifies the manifold structure across the feature space and the label space for label inference, but fails to align the geometric structures across different feature spaces \cite{DBLP:conf/cvpr/GongLCG19} to ensure smooth transfer learning between the source and target domains. Therefore, a hybridization of both \textbf{DGSA-GE-FE} and \textbf{DGSA-GE-LE} can be appealing for comprehensive manifold learning enhanced \textbf{DA} to leverage the advantages of both approaches.





In this paper, we propose a novel \textbf{DA} method, namely, Attention Regularized Laplace Graph-based Domain Adaptation (\textbf{ARG-DA}), to address the aforementioned '\textbf{Lack of Attention}' and '\textbf{Incomplete Unification}' issues. Specifically, the proposed (\textbf{ARG-DA}), \textbf{1)} optimizes the Attention Regularized Laplace Graph guided by \textbf{MMD}-based distribution measurements over cross-domain sub-tasks, thereby enforcing the attention regularized manifold learning and \textbf{2)} implements a hybridization strategy, namely, '\textbf{FEEL}', to unify the manifold structure across the feature/label spaces for comprehensive manifold alignment.

To conclude, the contributions of this paper are summarized as follows:

\begin{itemize}
	\item The Attention Regularized Laplace Graph is designed to enable discriminative manifold embedding, guided by \textbf{MMD}-based distribution measurements over cross-domain sub-tasks, thereby reducing negative transfer.

	\item A comprehensive manifold unification strategy, namely, \textbf{FEEL}, is proposed for comprehensive  manifold learning, which aligns the manifold structure across different feature spaces to minimize term.3 of Eq.(\ref{eq:bound}), while enforcing the alignment across feature space and label space so as to reduce term.1 of Eq.(\ref{eq:bound}).

	

	
	\item We carry out extensive experiments on 37 image classification DA tasks through 7 popular \textbf{DA} benchmarks and verify the effectiveness of the proposed method, which consistently outperforms thirty-three state-of-the-art \textbf{DA} algorithms by a significant margin. Moreover, we also carry out an in-depth analysis of the proposed \textbf{DA} methods, in particular, \textit{w.r.t.} their hyper-parameters and convergence speed. In addition, using real data, we also provide insights into the proposed \textbf{DA} model by analyzing the robustness of our proposed \textbf{ARG} term and \textbf{FEEL} strategy.

\end{itemize}

The paper is organized as follows. Sect.\ref{Related Work} discusses the related work. Sect.\ref{The proposed method} presents the method. Sect.\ref{Experiments} benchmarks the proposed DA method and provides an in-depth analysis. Sect.\ref{Conclusion} draws the conclusion.

\begin{figure}[h!]
	\vspace{-1em}
	\centering
	\includegraphics[width=1\linewidth]{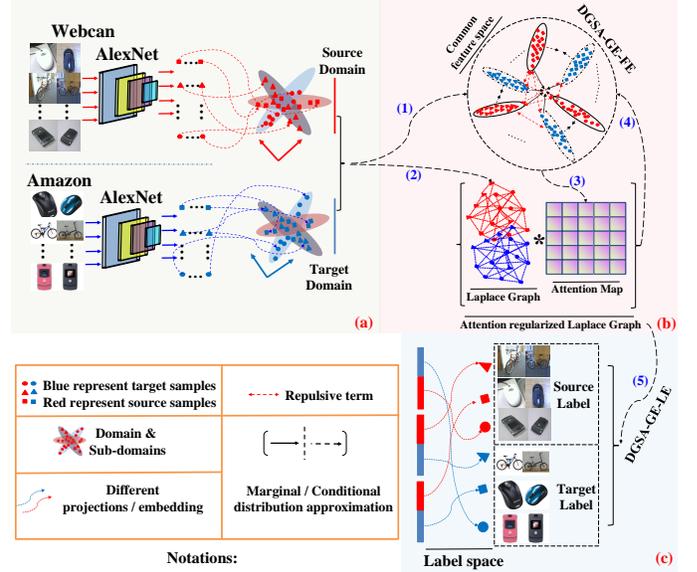}
	
	\caption {Illustration of the proposed \textbf{ARG-DA} method. Fig.\ref{fig:2} (a): source data and target data, \textit{e.g.},  mouse, bike, smartphone images, with different distributions and inherent hidden data geometric structures between the source in red and the target in blue. Samples of different class labels are represented by different geometric shapes, \textit{e.g.}, circle, triangle, and square; 	Fig.\ref{fig:2} (b).(1) illustrates \textbf{ARG-DA}, which optimizes data distributions closely yet distinctively by making use of the nonparametric distance, \textit{i.e.}, Maximum Mean Discrepancy (MMD); Fig.\ref{fig:2} (b).(2) designs the Laplace graph based on the original feature representation;  Fig.\ref{fig:2} (b).(3) calculates the Attention Map for weighting the varying degrees of importance across the differently labeled sub-domains; Fig.\ref{fig:2} (b).(4) aligns the manifold structure across different feature spaces using the Attention Regularized Laplace graph; Fig.\ref{fig:2} (b).(5) aligns the manifold structure across the feature space and the label space using the Attention Regularized Laplace Graph;  Fig.\ref{fig:2} (c): the achieved classification results in the discriminative label space.}
	\label{fig:2}
	\vspace{-1em}
\end{figure}


\vspace{-2mm}

\section{Related Work}
\label{Related Work}


State-of-the-art \textbf{DA} techniques feature two main research lines: 1) Shallow \textbf{DA}; 2) Deep \textbf{DA}. These are overviewed in Sect.\ref{subsect: Shallow DA} and Sect.\ref{subsect: Deep DA}, respectively, and discussed in comparison with the proposed \textbf{ARG-DA} simultaneously.


\vspace{-2mm}

\subsection{Shallow Domain Adaptation}
\label{subsect: Shallow DA}

\subsubsection{Geometric Alignment-based DA}
\label{Geometric alignment-based DA}

The rationale of geometric alignment-based \textbf{DA}  argues that the domain divergence can be reduced by aligning the hidden geometric structure across different domains, while the theoretical explanation is to minimize term.3 of Eq.(\ref{eq:bound}). Recent research in geometric alignment-based \textbf{DA} can be distinguished based on whether it incorporates the graph embedding techniques or embraces the subspace alignment approaches.


\textbf{Graph Embedding-based Domain Adaptation (DGSA-GE):}

By using the graph embedding techniques, \textbf{DGSA-GE} explicitly aligns the hidden manifold across different feature/label spaces for better function learning. To be specific, \textbf{DGSA-GE} can be categorized into \textbf{DGSA-GE-FE} and \textbf{DGSA-GE-LE} groups by aligning the manifold structure among different feature/label spaces.



\textbf{(a).} \textit{Feature Space Embedding-based DA (DGSA-GE-FE):}
\textbf{DGSA-GE-FE} methods \cite{jing2020adaptive,zhang2019manifold} aim to search the newly projected common feature space for cross-domain alignment through aligning the manifold structure in the original feature space. For instance, \textbf{ARTL} \cite{long2013adaptation} uses the \textbf{GE} strategy to align the different feature spaces and to optimize the \textbf{MMD} distance in the unified framework. Jingjing Li \textit{et al.}  \cite{li2018heterogeneous,li2018transfer} proposes cosine similarity to weight the diversity among different samples for local consistency preservation-based \textbf{DA}. To avoid the geometric structure distortion that existed in the original space, \textbf{MEDA} \cite{wang2018visual} specifically learns the Grassmann manifold to remedy the raised issue.  Different from traditional methods, \textbf{ACE} \cite{jing2020adaptive} argues that the previous research  significantly ignores the discriminative effectiveness, thus, proposes to improve the discriminativeness of the graph model by increasing the within-class distance and reducing the between-class distance across different sub-domains. \textbf{LGA-DA}  \cite{zhang2018structural} assumes that the source and target domains can share the same separation structure and ensures effective \textbf{DA}, by aligning their corresponding Laplace graphs' eigenspaces. In \textbf{RHGM}  \cite{das2018unsupervised}, hyper-graphs are used to find matches between samples of the source and target domains using similarities of different orders between graphs and achieve a hyper-graph matching based \textbf{DA}.

However, despite huge progress achieved by \textbf{DGSA-GE-FE} methods, \textbf{DGSA-GE-FE} methods only make use of the manifold alignment across different feature spaces to reduce term.3 of Eq.(\ref{eq:bound}), which fails to uniformly align the label space for a comprehensive  manifold  unification-based \textbf{DA}, thereby preventing it from harnessing the discriminative effectiveness of label space for shrinking of term.1 in Eq.(\ref{eq:bound}).


\textbf{(b).} \textit{Label Space Embedding-based DA (DGSA-GE-LE):}

Different from \textbf{DGSA-GE-FE}, \textbf{DGSA-GE-LE} explores better usage of the discriminative label space for proper functional learning on the source domain, thereby additionally reducing term.1 of Eq.(\ref{eq:bound}). For this purpose, \textbf{EDA} \cite{zhang2016robust} and  \textbf{DTLC} \cite{li2020discriminative} enforce the manifold structure alignment across the feature space and the label space, thereby generating the discriminative conditional distribution adaptation. By incorporating the label smoothness consistency (\textbf{LSmC}), \textbf{UDA-LSC} \cite{DBLP:journals/tip/HouTYW16} well accounts for the geometric structures of the underlying data manifold of the label space on the target domain, while \textbf{DGA-DA} \cite{luo2020discriminative} further improves \textbf{UDA-LSC} by additionally dealing with manifold regularization on the source domain simultaneously. \textbf{DAS-GA} \cite{pilanci2020domain} unifies the cross-domain graph bases to align the feature space and the label space for a spectrum transfer guided  \textbf{DA}.

Unfortunately, \textbf{DGSA-GE-LE} ignores the unification across different feature spaces as proposed in \textbf{DGSA-GE-FE}, thereby sacrificing smooth adaptation and resulting in incomplete  manifold unification. More importantly, both the \textbf{DGSA-GE-LE} and \textbf{DGSA-GE-FE} methods fail to weigh the importance across the different sub-tasks corresponding to different sub-domains for Attention Regularized \textbf{DA}, thereby increasing the risk of negative transfer.

\textbf{Subspace Alignment-based Domain Adaptation (DGSA-SA):}
Apart from \textbf{DGSA-GE}, an increasing number of \textbf{DA} methods, \textit{e.g.}, \cite{DBLP:journals/corr/LuoWHC17,DBLP:journals/ijcv/ShaoKF14,DBLP:journals/tip/XuFWLZ16,sun2016return,DBLP:journals/tnn/DingF18,DBLP:conf/iccv/FernandoHST13}, emphasize the importance of aligning the underlying data subspace rather than the graph model between the source and target domains for effective \textbf{DA}. In these methods, low-rank and sparse constraints are introduced into \textbf{DA} to extract a low-dimension feature subspace where target samples can be sparsely reconstructed from source samples \cite{DBLP:journals/ijcv/ShaoKF14}, or interleaved by source samples \cite{DBLP:journals/tip/XuFWLZ16}, thereby aligning the geometric structures of the underlying data manifolds. A few recent \textbf{DA} methods, \textit{e.g.}, \textbf{RSA-CDDA} \cite{DBLP:journals/corr/LuoWHC17}, \textbf{JGSA} \cite{Zhang_2017_CVPR}, further propose unified frameworks to reduce the shift between domains both statistically and geometrically. \textbf{HCA} \cite{liu2019homologous} improves \textbf{JGSA} using a homologous constraint on the two transformations for the source and target domains, respectively, to make the transformed domains related and hence alleviate negative domain adaptation. Despite the enormous progress achieved by \textbf{DGSA-SA}, we can see that merely aligning the data subspace across domains is also unable to ensure the theoretical effectiveness of \textbf{DGSA-SA} for shrinking the data distributions as required in term.2 of Eq.(\ref{eq:bound}).

Our proposed \textbf{ARG-DA} improves both \textbf{DGSA-GE} and \textbf{DGSA-SA} by jointly aligning the data distributions discriminatively and embracing the attention mechanism for Attention Regularized manifold learning-based \textbf{DA}. Moreover, \textbf{ARG-DA} embraces the proposed \textbf{FEEL} strategy to align the manifold structure across the whole feature/label spaces for the comprehensive manifold unification aware \textbf{DA}.

\subsubsection{Statistic Alignment-based DA \textbf{(STA-DA)}} 
\label{Statistic alignment-based DA}


The rationale of \textbf{STA-DA} is to assume that the existing domain divergence among the source and target domains can be significantly reduced by searching the newly optimized common feature space using the statistic measurements. Therefore, the learned knowledge from the source domain can be seamlessly applied to the target domain in the optimized common feature space.

For this purpose, recent research embraces a series of statistic measurements, \textit{e.g.}, Bregman Divergence \cite{4967588}, Wasserstein distance \cite{courty2017joint, courty2017optimal}, and \textit{Maximum Mean Discrepancy} (\textbf{MMD}) measurement \cite{pan2011domain}, to enforce the cross domain divergence reduced common feature space, thus subsequently reducing term.2 of Eq.(\ref{eq:bound}). Specifically, by embracing the dimensionality reduction techniques, \textbf{TCA} \cite{pan2011domain} explicitly minimizes the mismatch between source and target in terms of marginal distribution \cite{4967588} \cite{pan2011domain}. \textbf{JDA} \cite{long2013transfer} builds upon \textbf{TCA}'s framework and further exploits the conditional distribution alignment for effective sub-domain adaptation. Apart from previous \textbf{STA-DA} methods, \textbf{ILS} \cite{herath2017learning} learns the 
discriminative latent space using Mahalanobis metric to match statistical properties across different domains. Meanwhile,  \cite{lu2018embarrassingly} also explores the discriminative effectiveness as in \textbf{DA} by borrowing the merits of linear discriminant analysis (\textbf{LDA}) to optimize the common feature space.



However, despite this, great success has been achieved by \textbf{STA-DA} methods in reducing term.2 of Eq.(\ref{eq:bound}), which significantly fails to use manifold regularization for additionally dealing with  term.3 of Eq.(\ref{eq:bound}). Our proposed \textbf{ARG-DA} hybridizes the effectiveness of both \textbf{DGSA-GE} and \textbf{STA-DA}, and aims to develop an effective \textbf{DA} method to account for both the statistic and geometric aspects.

\vspace{-4mm}

\subsection{Deep Domain Adaptation}
\label{subsect: Deep DA}

Recently, deep learning techniques shed new light on \textbf{UDA} by borrowing the merits of high discriminative feature representations enabled by deep learning. They featured the following two main approaches.


\subsubsection{Statistic Matching-based DA}
\label{Statistic matching-based DA}

These approaches share similar spirits  to \textbf{STA-DA} by reducing the domain shift via different statistic measurements. However, this domain shift reduction is achieved by incorporating the \textbf{DL} paradigm rather than the shallow functional learning as proposed in \textbf{STA-DA}. The popular \textbf{DAN} \cite{long2015learning} reduces the marginal distribution divergence by incorporating the multi-kernel MMD loss on the fully connected layers of AlexNet. \textbf{JAN}  \cite{DBLP:conf/icml/LongZ0J17} improves \textbf{DAN} by jointly decreasing the divergence of both the marginal and conditional distributions.  \textbf{D-CORAL} \cite{sun2016deep} further introduces the second-order statistics into the AlexNet \cite{krizhevsky2012imagenet} framework for a more effective \textbf{DA} strategy. Different from previous research, \textbf{FDA} \cite{yang2020fda} and \textbf{MSDA} \cite{he2021multi} provide novel insight into shrinking the domain shift. \textbf{FDA} argues that the dissimilarities across domains can be measured by the low-frequency spectrum, while \textbf{MSDA} insists on aligning different domains by narrowing the \textbf{LAB} feature space representation, thereby swapping the contents of different low-frequency spectrum \cite{yang2020fda}, while \textbf{LAB} feature space representation \cite{he2021multi} among cross-domains enables qualified domain shift reduction. \textbf{ATM} \cite{li2020maximum} introduces an original distance loss, namely, maximum density divergence (\textbf{MDD}) to quantify distribution divergence for effective \textbf{DA}, and combines adversarial deep learning with metric learning. \textbf{BOS} \cite{jing2021balanced} projects the cross-domain samples onto a hyper-spherical latent space to implement the angular distance measured distribution alignment within the open set domain adaptation experimental scenarios. \textbf{CDCL} \cite{qian2015cross} proposes a generic cross-domain collaborative learning framework to collaboratively learn a shared feature representation for an effective cross-domain data analysis. \textbf{ETD} \cite{li2020enhanced} designed an enhanced transport distance to quantify the divergence between the cross-domain samples for  attention aware  \textbf{DA}. Both \textbf{SHOT} \cite{liang2020we} and \textbf{MA-UDA} \cite{li2020model} aim to solve a novel \textbf{DA} setting, namely, source-free domain adaptation (\textbf{SFDA}). \textbf{SHOT} hybridizes the mutual information and self-supervised pseudo-labeling techniques for effective \textbf{SFDA}, while \textbf{MA-UDA} develops a novel generative model to produce target-style training samples for  \textbf{SFDA}.


\subsubsection{Adversarial Loss-based DA}
\label{Adversarial loss-based DA}
These methods make use of \textbf{GAN} \cite{goodfellow2014generative} and propose to align data distributions across domains by making sample features indistinguishable \textit{w.r.t} the domain labels through an adversarial loss on a domain classifier \cite{ganin2016domain,tzeng2017adversarial,pei2018multi}. \textbf{DANN} \cite{ganin2016domain} and \textbf{ADDA} \cite{tzeng2017adversarial} learn a domain-invariant feature subspace by reducing the marginal distribution divergence. \textbf{MADA} \cite{pei2018multi} additionally makes use of multiple domain discriminators, thereby aligning conditional data distributions. Different from the previous approaches, \textbf{DSN} \cite{bousmalis2016domain} achieves domain-invariant representations by explicitly separating the similarities and dissimilarities in the source and target domains. \textbf{MADAN} \cite{zhao2019multi} explores knowledge from different multi-source domains to fulfill \textbf{DA} tasks. \textbf{CyCADA} \cite{pmlr-v80-hoffman18a} addresses the distribution divergence using a bi-directional \textbf{GAN}-based training framework. \textbf{UDA-GCN} \cite{wu2020unsupervised} facilitates knowledge transfer across domains based on the specifically designed dual graph convolutional network. \textbf{DAGNN} \cite{wu2019domain} designs the domain-adversarial graph neural network to comprehensively capture the semantic knowledge for effective text classification. \textbf{MDDA} \cite{zhang2020multimodal} hybridizes the multimodal disentangled representation learning and \textbf{UDA} techniques within a unified framework for prompt rumor detection of the newly emerged events.

The main advantage of these \textbf{DL}-based \textbf{DA} methods is that they jointly shrink the divergence of data distributions across domains and achieve a discriminative feature representation of data through a single unified end-to-end learning framework. Therefore, the optimized model naturally benefits from the merits similar to those in \textbf{META} learning \cite{wei2021metaalign}, which argues that the model can be reinforced by harnessing different tasks so as to receive the best candidate hyper-parameters. However, they generally suffer from the '\textit{batch learning}' \cite{goodfellow2016deep} strategy in contrast to '\textit{global modeling}' \cite{belkin2003laplacian} based optimization. As a result,  the manifold alignment of \textbf{DL}-based \textbf{DA} is generally implemented locally within the randomly selected batch samples rather than the global view enforced graph model \cite{Zhou04learningwith}. In this research work, our proposed \textbf{ARG-DA} is global manifold structure alignment-based optimization, while the rationale can also be easily explained, thereby enabling insights into further improving \textbf{DA} methods.



\vspace{-4mm}


\section{The proposed method}
\label{The proposed method}

In Sect.\ref{The proposed method}, we first define the notations. We set out the \textbf{DA} problem in Sect.\ref{Notations and Problem Statement}, after which Sect.\ref{subsection:Formulation} formulates our \textbf{DA} model. Sect.\ref{Solving the model} presents the method for solving the proposed \textbf{DA} model and derives the algorithm of \textbf{ARG-DA}. Sect.\ref{Kernelization} extends \textbf{ARG-DA} to non-linear problems through kernel mapping.

\vspace{-4mm}
\subsection{Notations and Problem Statement}
\label{Notations and Problem Statement}

Matrices are written as boldface uppercase letters. Vectors are written as boldface lowercase letters. For matrix ${\bf{M}} = ({m_{ij}})$, its $i$-th row is denoted by ${{\bf{m}}^i}$, and its $j$-th column by ${{\bf{m}}_j}$. We define the Frobenius norm ${\left\| . \right\|_F}$ as: ${\left\| {\bf{M}} \right\|_F} = \sqrt {\sum {_{i = 1}^n} \sum {_{j = 1}^l} m_{ij}^2} $. A domain $D$ is defined as an \textit{l}-dimensional feature space $\chi$ and a marginal probability distribution $P(x)$, \textit{i.e.}, $\mathcal{D}=\{\chi,P(x)\}$ with $x\in \chi$.  Given a specific domain $D$, a  task $T$ is composed of a \textit{C}-cardinality label set $\mathcal{Y}$  and a classifier $f(x)$,\textit{ i.e.}, $T = \{\mathcal{Y},f(x)\}$, where $f({x}) = \mathcal{Q}( y |x)$ can be interpreted as the class conditional probability distribution for each input sample $x$.

In unsupervised domain adaptation, we are given a source domain $\mathcal{D_S}=\{x_{i}^{s},y_{i}^{s}\}_{i=1}^{n_s}$ with $n_s$ labeled samples ${{\bf{X}}_{\cal S}} = [x_1^s...x_{{n_s}}^s]$, which are associated with their class labels ${{\bf{Y}}_S} = {\{ {y_1},...,{y_{{n_s}}}\} ^T} \in {{\bf{\mathbb{R}}}^{{n_s} \times C}}$, and an unlabeled target domain $\mathcal{D_T}=\{x_{j}^{t}\}_{j=1}^{n_t}$ with $n_t$  unlabeled samples ${{\bf{X}}_{\cal T}} = [x_1^t...x_{{n_t}}^t]$, whose labels  ${{\bf{Y}}_T} = {\{ {y_{{n_s} + 1}},...,{y_{{n_s} + {n_t}}}\} ^T} \in {{\bf{\mathbb{R}}}^{{n_t} \times C}}$ are unknown. Here, ${y_i} \in {{\bf{\mathbb{R}}}^c}(1 \le i \le {n_s} + {n_t})$ is a one-vs-all label hot vector in which $y_i^j = 1$ if ${x_i}$ belongs to the $j$-th class, and $0$ otherwise. We  define the data matrix ${\bf{X}} = [{{\bf{X}}_S},{{\bf{X}}_T}] \in {R^{l*n}}$ ($l$ = feature dimension; $n = {n_s} + {n_t}$ ) in packing both the source and target data. The source domain $\mathcal{D_S}$ and target domain $\mathcal{D_T}$ are assumed to be different, \textit{i.e.},  $\mathcal{\chi}_S=\mathcal{{\chi}_T}$, $\mathcal{Y_S}=\mathcal{Y_T}$, $\mathcal{P}(\mathcal{\chi_S}) \neq \mathcal{P}(\mathcal{\chi_T})$, $\mathcal{Q}(\mathcal{Y_S}|\mathcal{\chi_{S}}) \neq \mathcal{Q}(\mathcal{Y_T}|\mathcal{\chi_{T}})$. We also define the notion of \textit{sub-domain}, \textit{i.e.}, class,  denoted as ${\cal D}_{\cal S}^{(c)}$, representing the set of samples in ${{\cal D}_{\cal S}}$ with the class label $c$. It is worth noting that the definition of sub-domains in the target domain, namely ${\cal D}_{\cal T}^{(c)}$, requires a base classifier,\textit{ e.g.}, Nearest Neighbor (NN), to attribute  pseudo labels for the samples in ${{\cal D}_{\cal T}}$.


Maximum Mean Discrepancy (MMD) is an effective non-parametric distance-measurement that compares the distributions of two datasets by mapping the data into Reproducing Kernel Hilbert Space \cite{borgwardt2006integrating} (RKHS). Given two distributions $\mathcal{P}$ and $\mathcal{Q}$, the MMD between $\mathcal{P}$ and $\mathcal{Q}$ is defined as:
	\vspace{-0.5em}
\begin{equation}
	\label{eq:MMD}
	Dist(P,Q) = \parallel \frac{1}{n_1} \sum^{n_1}_{i=1} \phi(p_i) - \frac{1}{n_2} \sum^{n_2}_{i=1} \phi(q_i) \parallel_{\mathcal{H}}
\end{equation}

where $P=\{ p_1, \ldots, p_{n_1} \}$ and $Q = \{ q_1, \ldots, q_{n_2} \}$ are two random variable sets from distributions $\mathcal{P}$ and $\mathcal{Q}$, respectively, and $\mathcal{H}$ is a universal RKHS with the Reproducing Kernel Mapping $\phi$: $f(x) = \langle \phi(x), f \rangle$, $\phi: \mathcal{X} \to \mathcal{H}$.

\subsection{Formulation}
\label{subsection:Formulation}

As shown in Fig.\ref{fig:2}, our proposed model (Sect.\ref{The Final model}) starts from Close yet Discriminative Domain Adaptation \textbf{(CDDA)}  \cite{luo2017close,luo2020discriminative} (Sect.\ref{Discriminative statistic alignment}) to seek a common feature space in which both term.1 and term.2 of Eq.(\ref{eq:bound}) are minimized. Then, we propose in Sect.\ref{Attention Regularized Laplace Graph} the Attention Regularized Laplace Graph \textbf{(ARG)}-based \textbf{DA}, guided by similarity and discriminativeness when adapting each sub-domain in both the source and target domains, thereby improving the traditional Laplace graph. Sect.\ref{ARG-based Manifold Alignment} implements the proposed '\textbf{FEEL}' strategy and makes use of the proposed \textbf{ARG} to align the manifold structures across different feature/label spaces for comprehensive manifold unification, thus reducing term.1 and term.3 of Eq.(\ref{eq:bound}) simultaneously. By integrating \textbf{CDDA} and \textbf{ARG}-based manifold learning, Sect.\ref{The Final model} formalizes our final model \textbf{(ARG-DA)} and thereby optimizes at the same time the three terms of the right-hand of Eq.(\ref{eq:bound}).


\subsubsection{Discriminative Statistic Alignment}
\label{Discriminative statistic alignment}

Given the fact that the main issue significantly affecting model transferability is the cross-domain distribution divergence, a discriminative approach, \textit{i.e.}, \textbf{CDDA} \cite{luo2017close,luo2020discriminative}, is first implemented here to reduce the mismatch of the marginal distribution and the conditional distribution between the source and target domains.



\begin{figure}[h!]
	
	\centering
	\includegraphics[width=1\linewidth]{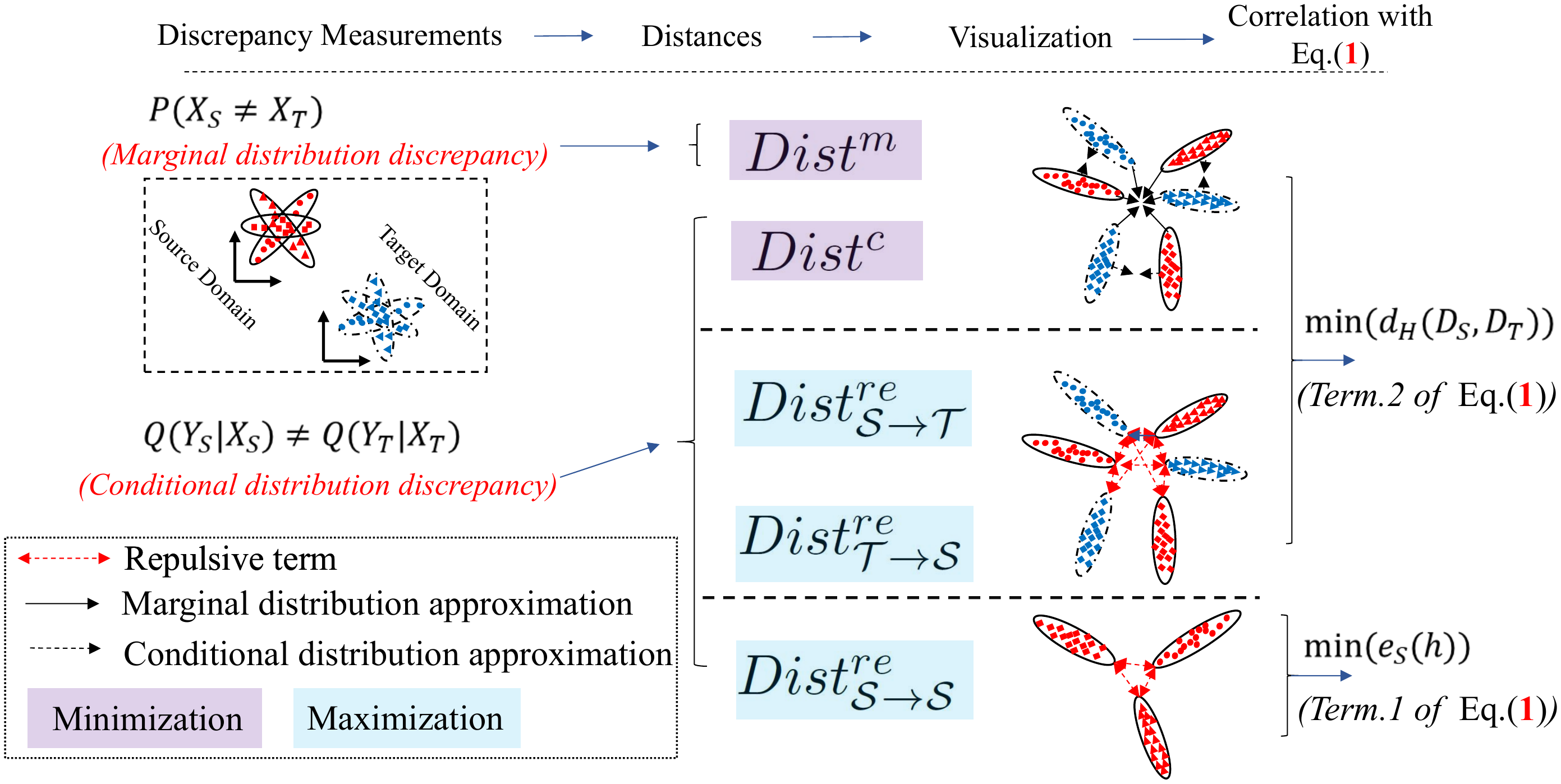}
	
	\caption {Purple parts represent distribution approximations across the source and target domains, \textit{w.r.t}, marginal and conditional distribution divergence minimization. Blue parts denote discriminative distribution approximations across and within the source and target domains, respectively.}

	\label{fig:3}
\end{figure} 

\textbf{Matching Marginal and Conditional Distributions}:
As shown in Fig.\ref{fig:3}, our model starts from \textbf{JDA}  \cite{long2013transfer}, which makes use of \textbf{MMD} in \textbf{RKHS} to measure the distances between the expectations of the source domain/sub-domain and target domain/sub-domain. Specifically, \textbf{1)} the empirical distance of the source and target domains is defined as $Dis{t^{m}}$; \textbf{2)} the conditional distance $Dis{t^{c}}$ is defined as the sum of the empirical distances between pairs of sub-domains in ${{\cal D}_{\cal S}}$ and ${{\cal D}_{\cal T}}$ with the same label; \textbf{3)} $Dis{t_{Clo}}$ is defined as the sum of $Dis{t^{m}}$ and $Dis{t^{c}}$, which will be optimized to bring closer both data distributions and class conditional data distributions across domains.
\begin{equation}\label{eq:JDA}
	\resizebox{0.7\hsize}{!}{%
		$\begin{array}{*{20}{l}}
		{Dis{t_{Clo}} = Dis{t^m}({D_S},{D_T}) + Dis{t^c}\sum\limits_{c = 1}^C {({D_S}^c,{D_T}^c)} } \\ 
		{\;\;\;\;\;\;\;\;\; = {{\left\| {\frac{1}{{{n_s}}}\sum\limits_{i = 1}^{{n_s}} {{{\mathbf{A}}^T}{x_i} - } \frac{1}{{{n_t}}}\sum\limits_{j = {n_s} + 1}^{{n_s} + {n_t}} {{{\mathbf{A}}^T}{x_j}} } \right\|}^2}} \\ 
		{\;\;\;\;\;\;\;\;\; + \sum\limits_{c = 1}^C {{{\left\| {\frac{1}{{n_s^{(c)}}}\sum\limits_{{x_i} \in {D_S}^{(c)}} {{{\mathbf{A}}^T}{x_i}}  - \frac{1}{{n_t^{(c)}}}\sum\limits_{{x_j} \in {D_T}^{(c)}} {{{\mathbf{A}}^T}{x_j}} } \right\|}^2}} } \\ 
		{\;\;\;\;\;\;\;\;\; = tr({{\mathbf{A}}^T}{\mathbf{X}}({{\mathbf{M}}_{\mathbf{0}}} + \sum\limits_{c = 1}^{c = C} {{{\mathbf{M}}_c}} ){{\mathbf{X}}^{\mathbf{T}}}{\mathbf{A}})} 
		\end{array}$}
\end{equation}

\begin{itemize}
	
	\item {$Dis{t^{m}}({{\cal D}_{\cal S}},{{\cal D}_{\cal T}})$}: where ${{\bf{M}}_0}$ is the MMD matrix  between ${{\cal D}_{\cal S}}$ and ${{\cal D}_{\cal T}}$ with ${{{({{\bf{M}}_0})}_{ij}} = \frac{1}{{{n_s}{n_s}}}}$ if $({{\bf{x}_i},{\bf{x}_j} \in {D_S}})$, ${{{({{\bf{M}}_0})}_{ij}} = \frac{1}{{{n_t}{n_t}}}}$ if $({{\bf{x}_i},{\bf{x}_j} \in {D_T}})$ and ${({{\bf{M}}_0})_{ij}} = \frac{{ - 1}}{{{n_s}{n_t}}}$ otherwise.  Thus, the difference between the marginal distributions $\mathcal{P}(\mathcal{X_S})$ and $\mathcal{P}(\mathcal{X_T})$ is reduced when minimizing {$Dis{t^{m}}({{\cal D}_{\cal S}},{{\cal D}_{\cal T}})$}.

	
	\item {$Dis{t^{c}}({{\cal D}_{\cal S}},{{\cal D}_{\cal T}})$}: where $C$ is the number of classes, $\mathcal{D_S}^{(c)} = \{ {\bf{x}_i}:{\bf{x}_i} \in \mathcal{D_S} \wedge y({\bf{x}_i}) = c\} $ represents the ${c^{th}}$ sub-domain in the source domain, in which $n_s^{(c)} = {\left\| {\mathcal{D_S}^{(c)}} \right\|_0}$ is the number of samples in the ${c^{th}}$ {source} sub-domain. $\mathcal{D_T}^{(c)}$ and $n_t^{(c)}$ are defined similarly for the target domain but using pseudo labels.  Finally, $\bf M_c$ denotes the MMD matrix between the sub-domains with labels $c$ in ${{\cal D}_{\cal S}}$ and ${{\cal D}_{\cal T}}$ with ${{{({{\bf{M}}_c})}_{ij}} = \frac{1}{{n_s^{(c)}n_s^{(c)}}}}$ if $({{\bf{x}_i},{\bf{x}_j} \in {D_S}^{(c)}})$, ${{{({{\bf{M}}_c})}_{ij}} = \frac{1}{{n_t^{(c)}n_t^{(c)}}}}$ if $({{\bf{x}_i},{\bf{x}_j} \in {D_T}^{(c)}})$, ${{{({{\bf{M}}_c})}_{ij}} = \frac{{ - 1}}{{n_s^{(c)}n_t^{(c)}}}}$ if $({{\bf{x}_i} \in {D_S}^{(c)},{\bf{x}_j} \in {D_T}^{(c)}\;or\;\;{\bf{x}_i} \in {D_T}^{(c)},{\bf{x}_j} \in {D_S}^{(c)}})$ and ${{{({{\bf{M}}_c})}_{ij}} = 0}$ otherwise. Consequently, the mismatch of conditional distributions between ${{D_{\cal S}}^c}$ and ${{D_{\cal T}}^c}$ is reduced by minimizing ${Dis{t^{c}}}$. 	
	
	\item Finally, the original data $\boldsymbol{X}$ are projected into the optimal common feature space using the mapping ${\bf{A}}$ through ${{\bf{A}}^T}{\bf{X}}$. 	
	
\end{itemize}

\textbf{Repulsive Force (RF) Across/Within Domains}: As shown in Fig.\ref{fig:3}, \textbf{JDA} is merely concerned with shrinking the \textbf{MMD} distances in order to reduce the cross-domain distribution divergence and ignores discriminative knowledge within data. Therefore, we directly borrow the idea from \textbf{CDDA} and \textbf{DGA-DA} \cite{luo2017close,luo2020discriminative} and introduce the \textit{Repulsive Force}(\textbf{RF}) term $Dis{t^{re}}{\text{ = }}Dist_{S \to T}^{re} + Dist_{T \to S}^{re}$  to further explore the discriminativeness of data. While \textbf{JDA} and \textbf{DGA-DA} have so far endeavored to minimize term.2 of Eq.(\ref{eq:bound}), in this paper we additionally optimize term.1 of Eq.(\ref{eq:bound}) as shown in Fig.\ref{fig:3} and introduce a novel \textbf{RF} term $Dist_{{\cal S} \to {\cal S}}^{re}$ as in Fig.\ref{fig:3}, so as to increase the discriminative power of the labeled source domain data, thereby allowing a better predictive model on the source domain and a decreasing term.1 of Eq.(\ref{eq:bound}). Subsequently, we use  ${{\cal S} \to {\cal T}}$, ${{\cal T} \to {\cal S}}$ and ${{\cal S} \to {\cal S}}$ to index the distances computed from ${D_{\cal S}}$ to ${D_{\cal T}}$,  ${D_{\cal T}}$ to ${D_{\cal S}}$ and ${D_{\cal S}}$ to ${D_{\cal S}}$, respectively, and $Dist_{{\cal S} \to {\cal T}}^{re}$ as the sum of the distances between each source sub-domain ${D_{\cal S}}^{(c)}$ and all the target sub-domains ${D_{\cal T}}^{(r);\;r \in \{ \{ 1...C\}  - \{ c\} \} }$ excluding the $c$-th target sub-domain. Symmetrically, $Dist_{{\cal T} \to {\cal S}}^{re}$ and $Dist_{{\cal S} \to {\cal S}}^{re}$ are defined similarly to $Dist_{{\cal S} \to {\cal T}}^{re}$. Consequently, the final Repulsive Force term $Dis{t^{re}}$ is formalized as:

\begin{equation}\label{eq:CDDAnew}
	\resizebox{0.9\hsize}{!}{$\begin{gathered}
		Dist_{S \to T}^{re} + Dist_{T \to S}^{re} + Dist_{S \to S}^{re} = Dis{t^c}\sum\limits_{c = 1}^C {({D_S}^c,{D_T}^{r \in \{ \{ 1...C\}  - \{ c\} \} })}  \hfill \\
		+ Dis{t^c}\sum\limits_{c = 1}^C {({D_T}^c,{D_S}^{r \in \{ \{ 1...C\}  - \{ c\} \} })}  + Dis{t^c}\sum\limits_{c = 1}^C {({D_S}^c,{D_S}^{r \in \{ \{ 1...C\}  - \{ c\} \} })}  \hfill \\
		= \sum\limits_{c = 1}^C {tr({{\mathbf{A}}^T}{\mathbf{X}}({{\mathbf{M}}_{S \to T}} + {{\mathbf{M}}_{T \to S}}{\text{    +    }}{{\mathbf{M}}_{S \to S}}){{\mathbf{X}}^{\mathbf{T}}}{\mathbf{A}})}  \hfill \\ 
		\end{gathered}$}
\end{equation}

Where
\begin{itemize}
	\item {${{\bf{M}}_{{\cal S} \to {\cal T}}}$ is defined as: ${{{({{\bf{M}}_{{\bf{S}} \to {\bf{T}}}})}_{ij}} = \frac{1}{{n_s^{(c)}n_s^{(c)}}}}$ if $({{\bf{x}_i},{\bf{x}_j} \in {D_S}^{(c)}})$, ${  \frac{1}{{n_t^{(r)}n_t^{(r)}}}}$ if $({{\bf{x}_i},{\bf{x}_j} \in {D_T}^{(r)}})$, ${  \frac{{ - 1}}{{n_s^{(c)}n_t^{(r)}}}}$ if $({{\bf{x}_i} \in {D_S}^{(c)},{\bf{x}_j} \in {D_T}^{(r)}\;or\;{\bf{x}_i} \in {D_T}^{(r)},{\bf{x}_j} \in {D_S}^{(c)}})$ and ${  0}$ otherwise.}
	
	\item {${{\bf{M}}_{{\cal T} \to {\cal S}}}$ is defined as: ${{{({{\bf{M}}_{{\bf{T}} \to {\bf{S}}}})}_{ij}} = \frac{1}{{n_t^{(c)}n_t^{(c)}}}}$ if ${({\bf{x}_i},{\bf{x}_j} \in {D_T}^{(c)})}$, ${  \frac{1}{{n_s^{(r)}n_s^{(r)}}}}$ if ${({\bf{x}_i},{\bf{x}_j} \in {D_S}^{(r)})}$, ${  \frac{{ - 1}}{{n_t^{(c)}n_s^{(r)}}}}$ if ${({\bf{x}_i} \in {D_T}^{(c)},{\bf{x}_j} \in {D_S}^{(r)}\;or\;{\bf{x}_i} \in {D_S}^{(r)},{\bf{x}_j} \in {D_T}^{(c)})}$ and ${  0}$ otherwise.}
	
	\item {${{\bf{M}}_{{\cal S} \to {\cal S}}}$ is defined as: ${{{({{\bf{M}}_{{\bf{S}} \to {\bf{S}}}})}_{ij}} = \frac{1}{{n_s^{(c)}n_s^{(c)}}}}$ if ${({x_i},{x_j} \in {D_S}^{(c)})}$, ${ \frac{1}{{n_s^{(r)}n_s^{(r)}}}}$ if ${({x_i},{x_j} \in {D_S}^{(r)})}$, ${  \frac{{ - 1}}{{n_s^{(c)}n_s^{(r)}}}}$ if ${({x_i} \in {D_S}^{(c)},{x_j} \in {D_S}^{(r)}\;or\;{x_i} \in {D_S}^{(r)},{x_j} \in {D_S}^{(c)})}$ and ${ 0}$ otherwise.}
	
\end{itemize}

Therefore, maximizing $Dist_{{\cal S} \to {\cal T}}^{re}$ and $Dist_{{\cal T} \to {\cal S}}^{re}$ increases the distances of each sub-domain with the other remaining sub-domains across domains, \textit{i.e.}, the between-class distances across domains, thereby reducing term.2 of Eq.(\ref{eq:bound}). On the other hand, maximizing $Dist_{{\cal S} \to {\cal S}}^{re}$ increases the between-class distances in the source domain, thereby optimizing term.1 in Eq.(\ref{eq:bound}) on classification errors on the source domain. In our model, the \textbf{RF} term is proposed within the source domain and across the different domains simultaneously for optimizing the first and second terms of the right-hand in Eq.(\ref{eq:bound})  simultaneously.


\subsubsection{Attention Regularized Laplace Graph}
\label{Attention Regularized Laplace Graph}

Although Sect.\ref{Discriminative statistic alignment} improves \textbf{CDDA} and \textbf{DGA-DA}  to optimize Term.1 and Term.2 of Eq.(\ref{eq:bound}), minimization of Term.3 in Eq.(\ref{eq:bound}) is not addressed and further requires manifold regularization in order to decrease the divergence between the classifiers on the source and target domains. However, as discussed in Sect.\ref{Introduction}, traditional manifold learning-based \textbf{DA} is not concerned with data class information and suffers from the so-called issue of '\textbf{Lack of Attention}'. To remedy this issue, we introduce here an Attention Regularized Laplace Graph (\textbf{ARG}) term to explicitly address the issues of \textbf{P1} and \textbf{P2} as highlighted in Sect.\ref{Introduction} and to achieve discriminativeness aware manifold learning. Specifically, the proposed \textbf{ARG} aims to achieve the following two goals:

\begin{itemize}
	\item \textit{Goal.1:}  To avoid the \textit{injudicious separation} of data from the same sub-domain as shown in Fig.\ref{fig:4}.(b), the proposed \textbf{ARG} needs to bring closer those cross sub-domains with the same label but with large domain divergence.


	\item \textit{Goal.2:} To prevent the \textit{brute force data geometric structure alignment} as highlighted in Fig.\ref{fig:4}.(d), the designed \textbf{ARG} needs to enable separation of those differently labeled yet closely aligned sub-domains.

	
\end{itemize}




\textbf{Mathematical Formalization:}
To meet \textit{Goal.1} and \textit{Goal.2}, the \textbf{ARG} term is mathematically formalized through the following steps:






\begin{figure}[h!]
	\vspace{-1em}
	\centering
	\includegraphics[width=1\linewidth]{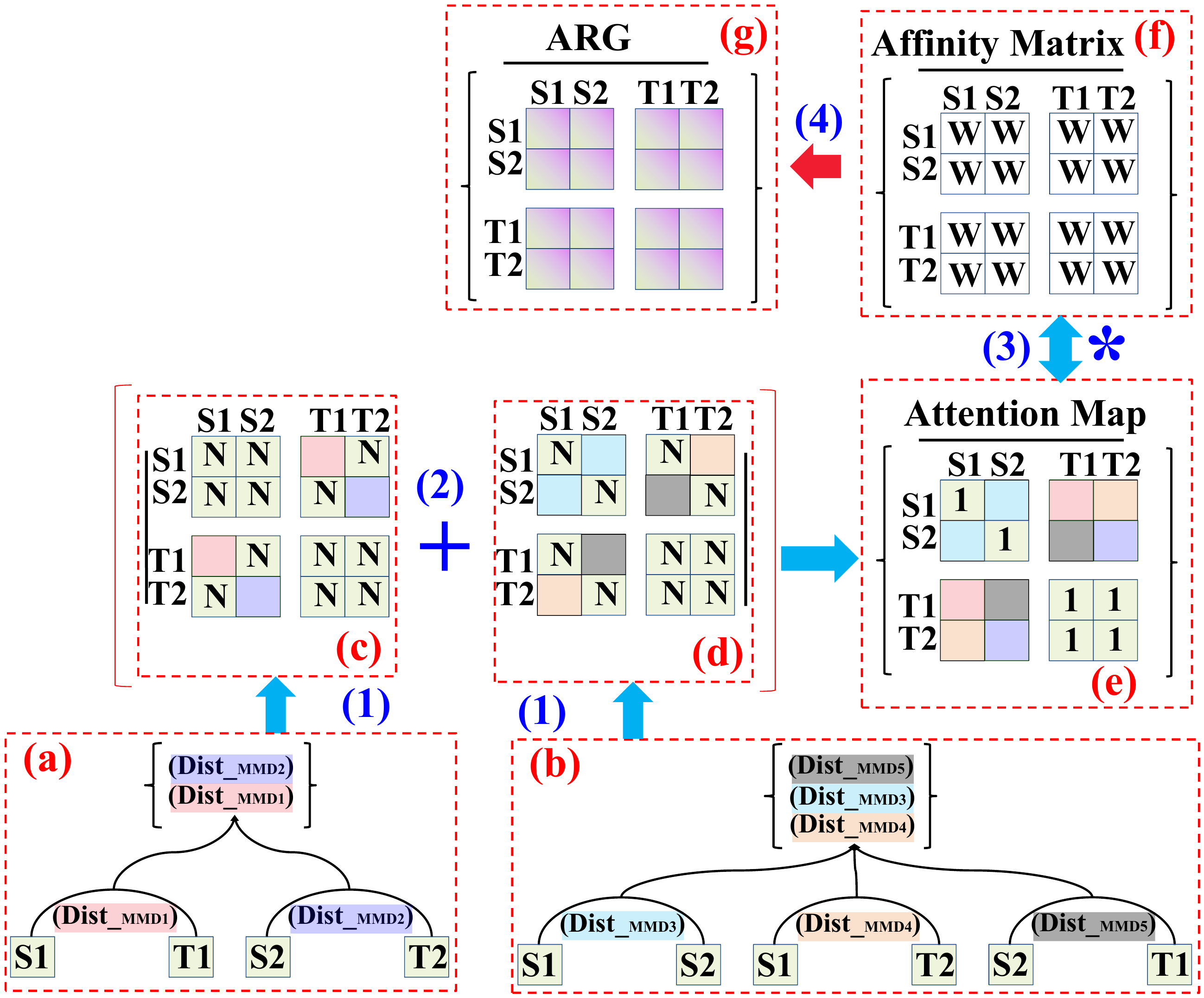}
	\caption {Detailed illustration of the proposed \textbf{A}ttention \textbf{R}egularized Laplace \textbf{G}raph. Fig.\ref{fig:5}.(a,b). Cross sub-domain divergence; 	Fig.\ref{fig:5}.(c). Intra-Class Attraction Aware Attention Map; 	Fig.\ref{fig:5}.(d). Inter-Class Repulsion Aware Attention Map; Fig.\ref{fig:5}.(e). Attention Map; Fig.\ref{fig:5}.(g). \textbf{A}ttention \textbf{R}egularized Laplace \textbf{G}raph.	} 
	\label{fig:5}
\end{figure}

\begin{itemize}
	
	\item  \textbf{Step.1}  \textit{- Design of an Affinity Matrix}: Initially, an affinity matrix ${\bf{W}}$ is defined to capture the relative relationship of each sample across the whole database, which is also the fundamental framework of the \textbf{ARG} term. Formally, we denote the pair-wise affinity matrix as:
	

	\begin{equation}\label{eq:5}
		\resizebox{0.82\hsize}{!}{${{\mathbf{W}}_{ij = }}\left\{ {\begin{array}{*{20}{c}}
				{sim({x_i},{x_j}),\;\;{x_i} \in {N_p}({x_j})\;or\;{x_j} \in {N_p}({x_i})} \\ 
				{0,\;\;\;\;\;\;\;\;\;\;\;\;\;\;\;otherwise\;\;\;\;\;\;\;\;\;\;\;\;\;\;\;\;\;\;\;\;\;\;\;} 
				\end{array}} \right.$}
	\end{equation}

	where  ${\bf{W}} = {[{w_{ij}}]_{({n_s} + {n_t}) \times ({n_s} + {n_t})}}$ is a symmetric matrix \cite{NIPS2001_2092}, with   ${w_{ij}}$ giving the affinity between two data samples $i$ and $j$ and defined as ${sim({x_i},{x_j})} = \exp ( - \frac{{{{\left\| {{\bf{x}_i} - {\bf{x}_j}} \right\|}^2}}}{{2{\sigma ^2}}})$ if $i \ne j$ and ${w_{ii}} = 0$ otherwise. Similar to \textbf{DGA-DA}, the greater the distance between a pair of samples ${dist({x_i},{x_j})}$, the less their similarity ${sim({x_i},{x_j})}$, and vice versa.

	
	\item  \textbf{Step.2}  \textit{- Computation of Cross Domain Intra-class Conditional Distribution Divergence}: given the labels ranging from $1 \to C$, leading to \textbf{C} pairs of sub-domains ${({D_S}^c,{D_T}^c)_{c \in (1...C)}}$ over the source and target domains, the cross domain intra sub-domain divergence is computed  as $Dist_{c \in (1...C)}^c({D_S}^c,{D_T}^c)$, using the \textbf{MMD} measurement:

	
	\begin{equation}\label{eq:6}
		\resizebox{0.82\hsize}{!}{$Dist_{c \in (1...C)}^c({D_S}^c,{D_T}^c) = {\left\| {\frac{1}{{n_s^{(c)}}}\sum\limits_{{x_i} \in {D_S}^{(c)}} {{{\mathbf{A}}^T}{x_i}}  - \frac{1}{{n_t^{(c)}}}\sum\limits_{{x_j} \in {D_T}^{(c)}} {{{\mathbf{A}}^T}{x_j}} } \right\|^2}$}
	\end{equation}
	
	Using Eq.(\ref{eq:6}),  \textbf{C}  cross sub-domain distances can be computed as illustrated in Fig.\ref{fig:5}.(a).
	
	
	\item  \textbf{Step.3}  \textit{(Intra-Class Attraction Aware \textbf{Attention Map})}: to meet \textit{Goal.1}, we aim to decrease the dissimilarities of sub-domain pairs of the same label but with large distances, thereby bringing closer these cross sub-domains for class aware data geometric alignment. For this purpose, as illustrated in Fig.\ref{fig:5}.(c), we define the Intra-Class Attraction Aware Attention (\textbf{A.Atten}) as:	


	\begin{equation}\label{eq:7}
		\resizebox{0.82\hsize}{!}{$\begin{gathered}
			A.Atten_{c \in (1...C)}^c({D_S}^c,{D_T}^c) \hfill \\
			= \frac{{Dist({D_S}^c,{D_T}^c) - \min (Dist_{c \in (1...C)}^c({D_S}^c,{D_T}^c)}}{{\max (Dist_{c \in (1...C)}^c({D_S}^c,{D_T}^c)) - \min (Dist_{c \in (1...C)}^c({D_S}^c,{D_T}^c))}} \hfill \\ 
			\end{gathered} $}
	\end{equation}

	By dot multiplication of Eq.(\ref{eq:5}) and Eq.(\ref{eq:7}), the affinity matrix receives an attention regularization, ${{\mathbf{W}}_{A.Atten}}{\text{ = }}{\mathbf{W}} \bullet A.Atten$, which amplifies the similarity of cross-domain data pairs when they belong to the same sub-domain. 

	\item  \textbf{Step.4}  \textit{(\textbf{RF} - Inter-class Conditional Distribution Divergence)}: We also calculate the conditional distribution divergence between differently labeled sub-domains in the source domain using \textbf{MMD} measurement.

	
	\begin{equation}\label{eq:8}
		\resizebox{0.6\hsize}{!}{$\begin{array}{l}
RF.Dist_{\left\{ {i,c} \right\} \in \left\{ {1...C} \right\};i \ne c}^{c,i}({D_S}^c,{D_S}^i)\\
 = {\left\| {\frac{1}{{n_s^{(c)}}}\sum\limits_{{x_i} \in {D_S}^{(c)}} {{{\bf{A}}^T}{x_i}}  - \frac{1}{{n_t^{(i)}}}\sum\limits_{{x_j} \in {D_S}^{(i)}} {{{\bf{A}}^T}{x_j}} } \right\|^2}
\end{array}$}
	\end{equation}
	
	Similar to Eq.(\ref{eq:6}), Eq.(\ref{eq:8}) captures the distance between the differently labeled sub-domains in the source domain in terms of conditional distributions. Inspired by the \textit{Repulsive Force}(\textbf{RF}) term as defined in Eq.(\ref{eq:CDDAnew}), we name the distance of Eq.(\ref{eq:8}) as $RF.Dist$ to distinguish it from the class attraction distance in Eq.(\ref{eq:6}).
	

	\begin{equation}\label{eq:9}
		\resizebox{0.6\hsize}{!}{$\begin{array}{l}
RF.Dist_{\left\{ {i,c} \right\} \in \left\{ {1...C} \right\};i \ne c}^{c,i}({D_S}^c,{D_T}^i)\\
 = {\left\| {\frac{1}{{n_s^{(c)}}}\sum\limits_{{x_i} \in {D_S}^{(c)}} {{{\bf{A}}^T}{x_i}}  - \frac{1}{{n_t^{(i)}}}\sum\limits_{{x_j} \in {D_T}^{(i)}} {{{\bf{A}}^T}{x_j}} } \right\|^2}
\end{array}$}
	\end{equation}
	
	As shown in Fig.\ref{fig:5}.(b), we also propose Eq.(\ref{eq:9}) to capture the distance between the differently labeled sub-domains across the source and target domains. 
	
	\item  \textbf{Step.5}  \textit{(Inter-Class Repulsion Aware \textbf{Attention Map})} To meet \textit{Goal.2}, we need to decrease the similarities of closely confounded yet differently labeled sub-domain data. As shown in Fig.\ref{fig:5}.(d), Inter-Class Repulsion Aware Attention (\textbf{R.Atten}) is defined as:
	

	\begin{equation}\label{eq:10}
		\resizebox{0.81\hsize}{!}{$\begin{gathered}
			R.Atten_{\left\{ {i,c} \right\} \in (1...C);i \ne c}^{c,i}({D_S}^c,\{ {D_S}^i,{D_T}^i\} ) =  \hfill \\
			\frac{{RF.Dist({D_S}^c,\{ {D_S}^i,{D_T}^i\} ) - \min (RF.Dist({D_S}^c,\{ {D_S}^i,{D_T}^i\} ))}}{{\max (RF.Dist({D_S}^c,\{ {D_S}^i,{D_T}^i\} )) - \min (RF.Dist({D_S}^c,\{ {D_S}^i,{D_T}^i\} ))}} \hfill \\ 
			\end{gathered} $}
	\end{equation}
	
	Using Eq.(\ref{eq:10}), the affinity matrix can be improved by attention regularization again, using the Inter-Class Repulsion Aware Attention Map (\textbf{R.AM}), through dot multiplication ${{\mathbf{W}}_{R.Atten}}{\text{ = }}{\mathbf{W}} \bullet R.Atten$, which decreases the affinities of sub-domain data that are differently labeled but very confounded in terms of data distributions. 

	\item  \textbf{Step.6}  \textit{(Attention Map)}: As shown in Fig.\ref{fig:5}.(c), by  unifying the Intra-Class Attraction Attention Map \textbf{A.AM} and the Inter-Class Repulsion Attention Map \textbf{R.AM}, we obtain our final \textbf{A}ttention \textbf{M}ap (\textbf{AM}), where the empty position is embedded with '\textbf{1}' to preserve the original manifold structures.

	
	\item  \textbf{Step.7}  \textit{(Attention Regularized Laplace Graph)}: Finally, the previous Attention Map is used to improve the traditional Laplace graph (${\mathbf{L}}$) in order to obtain the Attention Regularized Laplace Graph (\textbf{ARG}), which is mathematically formulated as:
	
	\begin{equation}\label{eq:12}
		\resizebox{0.55\hsize}{!}{$\mathop {{\mathbf{ARG}}}\limits_{{{\mathbf{W}}_{AM}}{\text{   =   }}{\mathbf{W}} \bullet {\mathbf{AM}}}  = ({\mathbf{I}} - {\mathbf{D}}_{AM}^{ - \frac{1}{2}}{{\mathbf{W}}_{AM}}{\mathbf{D}}_{AM}^{ - \frac{1}{2}})$}
	\end{equation}
	
	where ${{\mathbf{D}}_{AM}} = diag\{ {d_{A{M_{11}}}}...{d_{A{M_{({n_s} + {n_t}),({n_s} + {n_t})}}}}\} $ is the degree matrix with ${d_{A{M_{ii}}}} = \sum\nolimits_j {{w_{A{M_{ij}}}}}  $. The proposed \textbf{ARG} not only enables preservation of data manifold structures by following the standard spectral graph theory, but also modulates it by taking into account intra-class attraction and inter-class repulsion over the source and target domains, thereby leading to discriminative attention aware manifold embedding.
	


	


	\end {itemize}
	
\begin{figure}[h!]
		\vspace{-1em}
		\centering
		\includegraphics[width=1\linewidth]{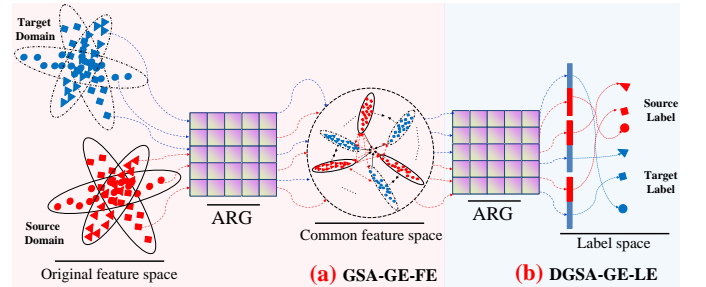}
		
		\caption {The proposed \textbf{FEEL} strategy is composed of \textbf{DGSA-GE-FE} and \textbf{DGSA-GE-LE}.} 
		\label{fig:6}
		\vspace{-1em}
	\end{figure}

	
	\subsubsection{\textbf{FEEL}-Based Manifold Alignment}
	\label{ARG-based Manifold Alignment}
	
	to deal with the second issue ('\textbf{Incomplete Unification}') as highlighted in Sect.\ref{Introduction}, we also propose a novel manifold learning strategy, which hybridizes the DGSA-GE-\textbf{FE} and DGSA-GE-\textbf{LE} approaches into a unified optimization framework and integrates seamlessly \textbf{ARG}-based embeddings, and formulate our final \textbf{FEEL} strategy for dynamic alignment of the manifold structures across different feature/label spaces as illustrated in Fig.\ref{fig:6}. Specifically, the proposed \textbf{FEEL} strategy aims to achieve the following three goals:

	\begin{itemize}
		\item \textit{Goal.3:}  Alignment of the manifold structures between the original feature space and the common feature space as enabled in the DGSA-GE-\textbf{FE} approach, thereby reducing term.3 of Eq.(\ref{eq:bound}).

		\item \textit{Goal.4:} Alignment of the common feature space and the label space as proposed in the DGSA-GE-\textbf{LE} techniques in order to decrease term.1 of Eq.(\ref{eq:bound}).
		
		
		\item \textit{Goal.5:} Seamless integration of the Attention Regularized Laplace Graph \textbf{ARG} for data distribution modulated graph-based  embeddings.


	\end{itemize}

	
	


	
	\textbf{Implementation details:} To meet \textit{Goal.3} through \textit{Goal.5}, \textbf{FEEL} hybridizes \textbf{(a)}. DGSA-GE-\textbf{FE} and \textbf{(b)}. DGSA-GE-\textbf{LE}. Furthermore, inspired by \textbf{DGA-DA} \cite{luo2020discriminative}, we also introduce the \textbf{(c)} Label Smoothness Consistency term \textbf{(LSmC)}, to ensure robust functional learning.
	
	\textbf{(a). Feature Space Embedding}: as shown in Fig.\ref{fig:6}.(a),  \textbf{FEEL} starts by searching a common feature space $\mathcal{(C)}$ from the original feature space $\mathcal{(X)}$  using matrix transformation ($\mathop {\mathbf{X}}\limits_{ \in \mathcal{X}}  \to \mathop {{\mathbf{AX}}}\limits_{ \in \mathcal{C}}$), while respecting the underlying data manifold geometric structure, and minimizes the following objective function:
	


	\begin{equation}\label{eq:13}
		\resizebox{0.75\hsize}{!}{$\begin{array}{l}
\min \sum\limits_{i,j} {{{({\bf{A}}{x_i} - {\bf{A}}{x_j})}^2}{\bf{W}}_{ij}^O} \\
 = \min (\sum\limits_{i,j} {({{({\bf{A}}{x_i})}^2} + {{({\bf{A}}{x_j})}^2} - 2{\bf{A}}{x_i}{\bf{A}}{x_j}){\bf{W}}_{ij}^O} ){\rm{ }}\\
 = \min (\sum\limits_i {{{({\bf{A}}{x_i})}^2}{\bf{D}}_{ii}^O}  + \sum\limits_j {{{({\bf{A}}{x_j})}^2}{\bf{D}}_{jj}^O}  - 2\sum\limits_{i,j} {{\bf{A}}{x_i}{\bf{A}}{x_j}{\bf{W}}_{ij}^O} ){\rm{ }}
\end{array}$}
	\end{equation}
	
	where ${{\mathbf{W}}^O}$ denotes the affinity matrix which is based on the \textbf{O}riginal feature space, and ${{\mathbf{D}}^O} = diag\{ d_{(1,1)}^O...d_{(({n_s} + {n_t}),({n_s} + {n_t}))}^O\} $ is the degree matrix with $d_{(i,i)}^O = \sum\nolimits_j {w_{(ij)}^O}  $. Optimizing Eq.(\ref{eq:13}) merely deals with \textit{Goal.3}. Therefore, we propose Eq.(\ref{eq:14}) to improve Eq.(\ref{eq:13}) via attention regularization to partially meet \textit{Goal.5}.


	
	\begin{equation}\label{eq:14}
		\resizebox{0.75\hsize}{!}{$\begin{array}{l}
\min ({{\bf{A}}^T}{\bf{XAR}}{{\bf{G}}^O}{{\bf{X}}^T}{\bf{A}})\\
s.t.\mathop {{\bf{AR}}{{\bf{G}}^O}}\limits_{{\bf{W}}_{AM}^O{\rm{     =     }}{{\bf{W}}^O} \bullet {\bf{AM}}}  = ({\bf{I}} - {({\bf{D}}_{AM}^O)^{ - \frac{1}{2}}}{\bf{W}}_{AM}^O{({\bf{D}}_{AM}^O)^{ - \frac{1}{2}}})
\end{array}$}
	\end{equation}
	
	Eq.(\ref{eq:14}) differs from Eq.(\ref{eq:13}) by reformulating ${{{\mathbf{W}}^O}}$ as ${{\mathbf{W}}_{AM}^O}$, thereby generating the attention aware DGSA-GE-\textbf{FE}.
	

	\textbf{(b). Label Space Embedding}: as shown in Fig.\ref{fig:6}.(b), in order to meet \textit{Goal.4} and \textit{Goal.5}, we further align the manifold structure across the common feature space $\mathcal{(C)}$ and the label space $\mathcal{(Y)}$ via ${\mathbf{AR}}{{\mathbf{G}}^C}$ regularized functional learning:


	\begin{equation}\label{eq:15}
		\resizebox{0.75\hsize}{!}{$\begin{gathered}
			\;\;\;\;\;\min ({{\mathbf{Y}}^T}{\mathbf{AR}}{{\mathbf{G}}^C}{\mathbf{Y}}) \hfill \\
			s.t.\mathop {{\mathbf{AR}}{{\mathbf{G}}^C}}\limits_{{\mathbf{W}}_{AM}^C{\text{     =     }}{{\mathbf{W}}^C} \bullet {\mathbf{AM}}}  = ({\mathbf{I}} - {({\mathbf{D}}_{AM}^C)^{ - \frac{1}{2}}}{\mathbf{W}}_{AM}^C{({\mathbf{D}}_{AM}^C)^{ - \frac{1}{2}}}) \hfill \\ 
			\end{gathered} $}
	\end{equation}
	
	in which  ${{\mathbf{W}}^C}$ is computed  in the \textbf{C}ommon feature space.

	\textbf{(c). Label Smoothness Consistency}: to enhance the robustness of dynamic functional learning across DGSA-GE-\textbf{FE} and DGSA-GE-\textbf{LE}, we borrow the merits of the label smoothness consistency \textbf{(LSmC)} term \cite{hou2016unsupervised, li2020discriminative} as proposed in \textbf{DGA-DA}:
	
	
	\begin{equation}\label{eq:16}
		\begin{array}{c}
			Dis{t^{lable}} = \sum\limits_{j = 1}^C {\mathop \sum \limits_{i = 1}^{{n_s} + {n_t}} } \left\| {\bf{Y}}^{(F)}_{i,j}-{\bf{Y}}^{(0)}_{i,j} \right\|
		\end{array}
	\end{equation}

	in which ${\bf{Y}} = {{\bf{Y}}_{\cal S}} \cup {{\bf{Y}}_{\cal T}}$,  ${\bf{Y}}_{i,j}^{(F)}$ denotes the probability of ${i_{th}}$ data belonging to ${j_{th}}$ class, and each piece of data ${x_i}$ has a predicted label ${y_i} = \arg {\max _{j \le c}}{\bf{Y}}_{ij}^F$. Indeed, the \textbf{(LSmC)} term is a constraint designed to prevent too many changes from the initial label assignment ${{\bf{Y}}_{\cal S}}$. By exploring the discriminative properties, we follow the experimental settings of \textbf{DGA-DA} and define the initial prediction ${{\bf{Y}}^{(0)}}$ as:
	
	\begin{equation}\label{eq:17}
		\resizebox{0.78\hsize}{!}{%
			$\begin{array}{*{20}{l}}
			{{\bf{Y}}_{{{\cal S}_{(ij)}}}^{(0)} = \left\{ {\begin{array}{*{20}{l}}
					{y_{{{\cal S}_{(ij)}}}^{(0)} = 1\;(1 \le i \le {n_s}),j = c,{y_{ij}} \in D_{\cal S}^{(c)}}\\
					{0\;\;\;\;\;\;\;\;\;\;\;else}
					\end{array}} \right.}\\
			{{\bf{Y}}_{{{\cal T}_{(ij)}}}^{(0)} = \left\{ {\begin{array}{*{20}{l}}
					\begin{array}{l}
					y_{{{\cal T}_{(ij)}}}^{(0)} = 1\;(({n_s} + 1) \le i \le {n_s} + {n_t}),j = c,\\
					{y_{ij}} \in D_{\cal T}^{(c)}
					\end{array}\\
					{0\;\;\;\;\;\;\;\;\;\;\;else}
					\end{array}} \right.}
			\end{array}$}
	\end{equation}
	
	where $D_{\cal T}^{(c)}$ is defined as pseudo labels, generated via a base classifier, \textit{e.g.}, \textbf{NN}. Therefore, jointly optimizing Eq.(\ref{eq:14}), Eq.(\ref{eq:15}) and Eq.(\ref{eq:16}) ensures \textit{Goal.4} through \textit{Goal.5} and guarantees robust optimization simultaneously.
	


	\subsubsection{The final model}
	\label{The Final model} 
	To sum up, our final model (\textbf{ARG-DA}) jointly optimizes Eq.(\ref{eq:JDA}), Eq.(\ref{eq:CDDAnew}), Eq.(\ref{eq:14}), Eq.(\ref{eq:15}), and Eq.(\ref{eq:16}) within a unified framework, which comprehensively minimizes the error bound as defined in Eq.(\ref{eq:bound})  and  simultaneously addresses the two issues highlighted in Sect.\ref{Introduction}, namely ('\textbf{Lack of Attention}' and '\textbf{Incomplete Unification}'). Specifically, it can be written mathematically as:

	
	\begin{equation}\label{eq:18}
		\resizebox{0.65\hsize}{!}{%
			$\begin{gathered}
			\min \left( {\begin{array}{*{20}{l}}
				{\sum\limits_{c = 0}^C {tr({{\mathbf{A}}^T}{\mathbf{X}}({{\mathbf{M}}^*} + {\mathbf{AR}}{{\mathbf{G}}^O}){{\mathbf{X}}^T}{\mathbf{A}})}  + \lambda \left\| {\mathbf{A}} \right\|_F^2} \\ 
				{ + \mu (\sum\limits_{j = 1}^C {\sum\limits_{i = 1}^{{n_s} + {n_t}} {\left\| {{\mathbf{Y}}_{ij}^{(F)} - {\mathbf{Y}}_{ij}^{(0)}} \right\|} } ) + {{\mathbf{Y}}^T}({\mathbf{AR}}{{\mathbf{G}}^C}){\mathbf{Y}}} 
				\end{array}} \right) \hfill \\
			{\text{s}}t.{{\mathbf{M}}^*} = {{\mathbf{M}}_0} + \sum\limits_{c = 1}^C {({{\mathbf{M}}_c})}  - ({{\mathbf{M}}_{S \to T}} + {{\mathbf{M}}_{T \to S}}{\text{  +  }}{{\mathbf{M}}_{S \to S}}), \hfill \\
			{{\mathbf{A}}^T}{\mathbf{XH}}{{\mathbf{X}}^T}{\mathbf{A}} = {\mathbf{I}},\;\mathop {{\mathbf{AR}}{{\mathbf{G}}^O}}\limits_{{\mathbf{W}}_{AM}^O{\text{     =     }}{{\mathbf{W}}^O} \bullet {\mathbf{AM}}} ,\;\mathop {{\mathbf{AR}}{{\mathbf{G}}^C}}\limits_{{\mathbf{W}}_{AM}^C{\text{      =      }}{{\mathbf{W}}^C} \bullet {\mathbf{AM}}}  \hfill \\ 
			\end{gathered} $}
	\end{equation}
	
	where ${{\mathbf{M}}^*}$ synthesizes the \textbf{MMD} matrices of both Eq.(\ref{eq:JDA}) and Eq.(\ref{eq:CDDAnew}) to ensure a discriminative statistic alignment as discussed in Sect.\ref{Discriminative statistic alignment}. The constraint ${{{\bf{A}}^T}{\bf{XH}}{{\bf{X}}^T}{\bf{A}} = {\bf{I}}}$ removes an arbitrary scaling factor in the embedding and prevents the above optimization from collapse onto a subspace of dimensions less than the required dimensions. On the other hand, $\lambda$ is a regularization parameter guaranteeing that the optimization problem is well-defined, and $\mu $ is a trade-off parameter which balances \textbf{LSmC} and label space embedding.
	
	Fig.\ref{fig:7} highlights the contribution of each component in Eq.(\ref{eq:18}) to decreasing the \textbf{three terms} of the error bound defined in Eq.(\ref{eq:bound}):

	\begin{figure}[h!]

		\centering
		\includegraphics[width=1\linewidth]{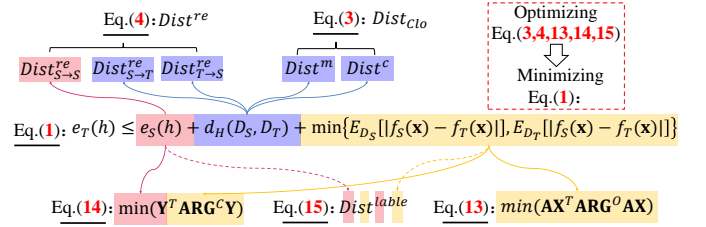}
		
		\caption {Highlight of Eq.(\ref{eq:18})  where the red, purple, and yellow parts denote the impact of various equations on Term.1, Term.2, and Term.3 in the error bound (Eq.(\ref{eq:bound})) limitation, respectively. } 
		\label{fig:7}
			\vspace{-1em}
			
	\end{figure}


	\begin{itemize}
		\item \textbf{Term.1}: leveraging the labels available in the source domain, both Eq.(\ref{eq:CDDAnew}) and Eq.(\ref{eq:15}) minimize  \textbf{Term.1} of Eq.(\ref{eq:bound}) via distribution alignment and manifold regularization, respectively, while Eq.(\ref{eq:16}) enforces minimization of \textbf{Term.1} through label propagation.


		\item \textbf{Term.2}: in line with the  statistic alignment approach, joint optimization of $Dis{t^m}+Dis{t^c}$ in Eq.(\ref{eq:JDA}) and $Dist_{S \to T}^{re} + Dist_{T \to S}^{re}$ in Eq.(\ref{eq:CDDAnew}) decreases the marginal and conditional distribution divergences, respectively, thereby enforcing minimization of \textbf{Term.2} of Eq.(\ref{eq:bound});

		\item \textbf{Term.3}: optimizing  Eq.(\ref{eq:14}) and Eq.(\ref{eq:15}) theoretically shrinks the divergence between different labeling functions, thereby reducing \textbf{Term.3} of Eq.(\ref{eq:bound}). Meanwhile, Eq.(\ref{eq:16}) partially shrinks  \textbf{Term.3} of Eq.(\ref{eq:bound}) thanks to the joint optimization of  Eq.(\ref{eq:15}) and Eq.(\ref{eq:16}).

		
	\end{itemize}


				\vspace{-1.5em}

	\subsection{Solving the model}
	\label{Solving the model}
	
	Optimization of our final model Eq.(\ref{eq:18}) is based on the following three key factors:

	\begin{itemize}
		\item \textbf{F.1}: Discriminative statistic alignment (Eq.(\ref{eq:JDA}), Eq.(\ref{eq:CDDAnew}));

		\item \textbf{F.2}: \textbf{ARG}-based DGSA-GE-\textbf{FE}, which aligns the manifold structure across different feature spaces, (Eq.(\ref{eq:14}));
		

		\item \textbf{F.3}: \textbf{ARG}-based DGSA-GE-\textbf{LE}, which aligns the manifold structure across feature and label spaces, (Eq.(\ref{eq:15}), Eq.(\ref{eq:16})).

		
	\end{itemize}

	These three key factors can be categorized into two sub-problems \textbf{(SPs)} for the dynamic functional learning:
	

	\begin{itemize}
		\item \textbf{SP.1}: in \textbf{SP.1}, both \textbf{F.1} and \textbf{F.2} enforce the domain alignment across the \textit{original feature space and the common feature space}. Mathematically, \textbf{SP.1} can be formalized as:


		\begin{equation}\label{eq:19}
			\resizebox{0.65\hsize}{!}{%
				$\begin{gathered}
				\min (\sum\limits_{c = 0}^C {tr({{\mathbf{A}}^T}{\mathbf{X}}({{\mathbf{M}}^*} + {\mathbf{AR}}{{\mathbf{G}}^O}){{\mathbf{X}}^T}{\mathbf{A}})}  + \lambda \left\| {\mathbf{A}} \right\|_F^2) \hfill \\
				{\text{s}}t.{{\mathbf{M}}^*} = {{\mathbf{M}}_0} + \sum\limits_{c = 1}^C {({{\mathbf{M}}_c})}  - ({{\mathbf{M}}_{S \to T}} + {{\mathbf{M}}_{T \to S}}{\text{   +   }}{{\mathbf{M}}_{S \to S}}), \hfill \\
				{{\mathbf{A}}^T}{\mathbf{XH}}{{\mathbf{X}}^T}{\mathbf{A}} = {\mathbf{I}},\;\mathop {{\mathbf{AR}}{{\mathbf{G}}^O}}\limits_{{\mathbf{W}}_{AM}^O{\text{      =      }}{{\mathbf{W}}^O} \bullet {\mathbf{AM}}}  \hfill \\ 
				\end{gathered} $}
		\end{equation}

		\item \textbf{SP.2}: using the updated feature representation ${{{\mathbf{A}}^T}{\mathbf{X}}}$ of \textbf{SP.1}, \textbf{SP.2} aligns the manifold structure across the \textit{common feature space and the label space} to achieve \textbf{F.3}.
		
		
		\begin{equation}\label{eq:20}
			\resizebox{0.65\hsize}{!}{%
				$\begin{gathered}
				\min (\mu (\sum\limits_{j = 1}^C {\sum\limits_{i = 1}^{{n_s} + {n_t}} {\left\| {{\mathbf{Y}}_{ij}^{(F)} - {\mathbf{Y}}_{ij}^{(0)}} \right\|} } ) + {{\mathbf{Y}}^T}({\mathbf{AR}}{{\mathbf{G}}^C}){\mathbf{Y}}) \hfill \\
				{\text{s}}t.\mathop {{\mathbf{AR}}{{\mathbf{G}}^C}}\limits_{{\mathbf{W}}_{AM}^C{\text{       =       }}{{\mathbf{W}}^C} \bullet {\mathbf{AM}}}  \hfill \\ 
				\end{gathered} $}
		\end{equation}
		
	\end{itemize}

	As a result, our final model Eq.(\ref{eq:18}) can thus be naturally divided into two sub-optimization problems (Eq.(\ref{eq:19}) and Eq.(\ref{eq:20})), each of which has a closed form solution and can be optimized alternatively. 
	
	Optimization of \textbf{SP.1}: Eq.(\ref{eq:19}) amounts to solving the generalized eigendecomposition problem in order to obtain the best matrix projection $\boldsymbol{A}$. By using the Augmented Lagrangian method  \cite{fortin2000augmented,long2013transfer}, we obtain the best candidate matrix projection through setting its partial derivation \textit{w.r.t.} $\boldsymbol{A}$ equal to zero:

		\begin{equation}\label{eq:21}
			\resizebox{0.6\hsize}{!}{%
				$({\mathbf{X}}({{\mathbf{M}}^*} + {\mathbf{AR}}{{\mathbf{G}}^O}){{\mathbf{X}}^T} + \lambda {\mathbf{I}}){\mathbf{A}} = {\mathbf{XH}}{{\mathbf{X}}^T}{\mathbf{A}}\Phi  $}
		\end{equation}

	where $\Phi {\rm{ = diagram}}({\varphi _1},...{\varphi _k}) \in {R^{k*k}}$ is the Lagrange multiplier, and ${\mathbf{AR}}{{\mathbf{G}}^O}$ is calculated using the original feature representation. The optimal subspace $\boldsymbol{A}$ is reduced to solving Eq.(\ref{eq:21}) for the k smallest eigenvectors. Then, we obtain the projection matrix $\boldsymbol{A}$ and the updated feature representation in the newly searched common feature space ${\bf{Z}} = {{\bf{A}}^T}{\bf{X}}$.

	Optimization of \textbf{SP.2}: using ${\bf{Z}}$ obtained in \textbf{SP.1}, we formulate ${\mathbf{AR}}{{\mathbf{G}}^C}$ to further explore the discriminative power of the label space ${\bf{Y}}$ by optimizing Eq.(\ref{eq:20}). Inspired by previous research \cite{Zhou04learningwith} \cite{6341755}, the minimum of Eq.(\ref{eq:20}) is approached where the derivative of the function is zero. Therefore, the approximate solution can be formulated as: 
	
			\begin{equation}\label{eq:22}
			\resizebox{0.5\hsize}{!}{%
				${{\mathbf{Y}}^ \star } = {({\mathbf{D}}_{AM}^C - (\frac{1}{{1 + \mu }}){\mathbf{W}}_{AM}^C)^{ - 1}}{{\mathbf{Y}}^{(0)}}$}
		\end{equation}

	where $\textbf{Y}^\star$ is the probability of prediction of the target domain corresponding to different class labels,  ${\mathbf{W}}_{AM}^C$ is the affinity matrix based on the feature representation in the optimized common feature space, and ${\mathbf{D}}_{AM}^C$ is the diagonal matrix.

	For simplicity's sake, we define $\alpha {\rm{ = }}\frac{1}{{1 + \mu }}$, then Eq.(\ref{eq:22}) is reformulated as Eq.(\ref{eq:23}):
	
	\begin{equation}\label{eq:23}
		{{\mathbf{Y}}^ \star } = {({\mathbf{D}}_{AM}^C - \alpha {\mathbf{W}}_{AM}^C)^{ - 1}}{{\mathbf{Y}}^{(0)}}
	\end{equation}

	
	To sum up, at a given iteration, \textbf{SP.1} and \textbf{SP.2} can be dynamically optimized to search the optimal solution and ensure the Attention Regularized Laplace Graph-based-\textbf{DA}.

	The complete learning algorithm is summarized in Algorithm 1 - \textbf{ARG-DA}.
	
	\begin{algorithm}[!h]
		\scriptsize
		\caption{Attention Regularized Laplace Graph for Domain Adaptation (\textbf{ARG-DA})}
		\KwIn{Data $\bf{X}$, Source domain labels ${\bf{Y}}_{\cal S}$, subspace dimension $k$, number of iterations $T$, regularization parameters $\lambda $ and $\alpha $}
		
		\textbf{Step {1}}:  Initialize the iteration counter t=0 and compute ${{\bf{M}}_0}$ as in Eq.(\ref{eq:JDA}).\\
		
		
		\textbf{Step 2}: Initialize pseudo target labels ${{\bf{Y}}_{\cal T}}$ and projection space $\textbf{A}$:\\
		
		\textbf{Step 3}:  Initialize pseudo target labels ${{\bf{Y}}_{\cal T}}$ via a base classifier, \textit{e.g.}, 1-NN, based on source domain labels ${{\bf{Y}}_{\cal S}}$.
		
		\textit{\textbf{Stage.1}:} 
		\While{ Definition of \textit{\textbf{Stage.1}:}}{
			(1) Initialize ${{\mathbf{AR}}{{\mathbf{G}}^O}}$ based on the pseudo target labels and the original feature representation $\bf{X}$ via Eq.(\ref{eq:12}).
			
			(2) Solve the generalized eigendecomposition problem \cite{fortin2000augmented,long2013transfer} as in Eq.(\ref{eq:21}) (replace ${\mathbf{M}}^*$ by ${{\bf{M}}_0}$ ) and obtain adaptation matrix $\bf A$, then  embed data via the transformation, $\bf{Z} = {{\bf{A}}^T}{\bf{X}}$\;
		}

		\textit{\textbf{Stage.2}:}
		\While{ Definition of \textit{\textbf{Stage.2}:}}{
			(1) Initialize ${{\mathbf{AR}}{{\mathbf{G}}^O}}$ based on the pseudo target labels and the common feature representation $\bf{Z}$ via Eq.(\ref{eq:12}).
			
			(2) Solve the generalized eigendecomposition problem \cite{fortin2000augmented,long2013transfer} as in Eq.(\ref{eq:21}) (replace ${\mathbf{M}}^*$ by ${{\bf{M}}_0}$ ) and obtain adaptation matrix $\bf A$, then  embed data via the transformation, $\bf{Z} = {{\bf{A}}^T}{\bf{X}}$\;
			
			(3) construct the label matrix ${{\bf{Y}}^{(0)}}$ as in Eq.(\ref{eq:17}), then obtain ${{\rm{{\bf{Y}}}}_{final}}$ by solving Eq.(\ref{eq:22})\;
			
			(4) update pseudo target labels $\{ {\bf{Y}}_{\cal T}^{(F)} = {{\bf{Y}}_{final}}\left[ {:,({n_s} + 1):({n_s} + {n_t})} \right]\} $;\\
		}

		\While{not converged and $t<T$	}{
			\textbf{Step 4}: Update projection space \textbf{A} \\
			
			(i) Compute ${\bf{M}}_c$ as in Eq.(\ref{eq:JDA}) 
			
			(ii) Compute ${\mathbf{M}}^*$ as defined in Eq.(\ref{eq:18}) via updated ${{\bf{Y}}_{\cal T}}$. 

			\textbf{Step 5}:	\textit{\textbf{Stage.1}:}	
			
			\textbf{Step 6}:    \textit{\textbf{Stage.2}:}
			
			\textbf{Step 7}: $t=t+1$; Return to Step 3; \\               
			
		}
		\KwOut{Adaptation matrix ${\bf{A}}$, embedding ${\bf{Z}}$, Target domain labels ${\bf{Y}}_{\cal T}^{(F)}$}
	\end{algorithm}

	\subsection{Kernelization}
	\label{Kernelization}
	The proposed \textbf{ARG-DA} method is extended to nonlinear problems in a Reproducing Kernel Hilbert Space via the kernel mapping $\phi :x \to \phi (x)$, or $\phi ({\bf{X}}):[\phi ({{\bf{x}}_1}),...,\phi ({{\bf{x}}_n})]$, and the kernel matrix ${\bf{K}} = \phi {({\bf{X}})^T}\phi ({\bf{X}}) \in {R^{n*n}}$. We utilize
	the representer theorem to formulate the Kernel \textbf{ARG-DA} as:
	
	\vspace{-0.8em}

	\begin{equation}\label{eq:24}
		\resizebox{0.7\hsize}{!}{%
			$\begin{gathered}
			\min \left( {\begin{array}{*{20}{l}}
				{\sum\limits_{c = 0}^C {tr({{\mathbf{A}}^T}{\mathbf{K}}({{\mathbf{M}}^*} + {\mathbf{AR}}{{\mathbf{G}}^O}){{\mathbf{K}}^T}{\mathbf{A}})}  + \lambda \left\| {\mathbf{A}} \right\|_F^2} \\ 
				{ + \mu (\sum\limits_{j = 1}^C {\sum\limits_{i = 1}^{{n_s} + {n_t}} {\left\| {{\mathbf{Y}}_{ij}^{(F)} - {\mathbf{Y}}_{ij}^{(0)}} \right\|} } ) + {{\mathbf{Y}}^T}({\mathbf{AR}}{{\mathbf{G}}^C}){\mathbf{Y}}} 
				\end{array}} \right) \hfill \\
			{\text{s}}t.{{\mathbf{M}}^*} = {{\mathbf{M}}_0} + \sum\limits_{c = 1}^C {({{\mathbf{M}}_c})}  - ({{\mathbf{M}}_{S \to T}} + {{\mathbf{M}}_{T \to S}}{\text{   +   }}{{\mathbf{M}}_{S \to S}}), \hfill \\
			{{\mathbf{A}}^T}{\mathbf{KH}}{{\mathbf{K}}^T}{\mathbf{A}} = {\mathbf{I}},\;\mathop {{\mathbf{AR}}{{\mathbf{G}}^O}}\limits_{{\mathbf{W}}_{AM}^O{\text{      =      }}{{\mathbf{W}}^O} \bullet {\mathbf{AM}}} ,\;\mathop {{\mathbf{AR}}{{\mathbf{G}}^C}}\limits_{{\mathbf{W}}_{AM}^C{\text{       =       }}{{\mathbf{W}}^C} \bullet {\mathbf{AM}}}  \hfill \\ 
			\end{gathered} $}
	\end{equation}

	\section{Experiments}
	\label{Experiments}
	
	\subsection{Benchmarks and Features}
	\label{subsection:Benchmarks and Features}
	
\begin{figure}[h!]

		\centering
		\includegraphics[width=1\linewidth]{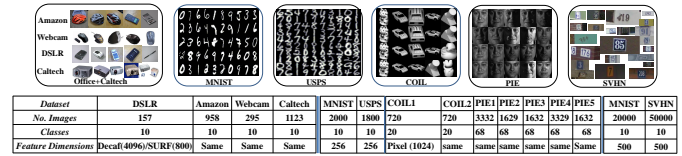}
		\caption { Sample images from seven datasets used in our experiments. Each dataset represents a different domain. The Office dataset in Office+Caltech contains three sub-datasets, namely DSLR, Amazon, and Webcam.} 
		\label{fig:8}	
		\vspace{-2em}
	\end{figure}

	In this paper, as shown in Fig.\ref{fig:8}, we evaluate the proposed \textbf{ARG-DA}  using standard benchmarks, \textit{i.e.}, USPS \cite{DBLP:journals/pami/Hull94}+MINIST \cite{lecun1998gradient}, COIL20 \cite{long2013transfer}, PIE \cite{long2013transfer}, office+Caltech \cite{long2013transfer}, and SVHN-MNIST \cite{bousmalis2016domain}, and compare it with state-of-the-art \textbf{DA} methods. Following the experimental settings as reported in the previous works \cite{DBLP:journals/pami/GhifaryBKZ17,DBLP:journals/tip/DingF17,DBLP:journals/corr/LuoWHC17,bousmalis2016domain,liang2018aggregating}, we construct 37 datasets for different image classification tasks.
	
	
	\textbf{Office+Caltech} consists of 2533 images of 10 categories (8 to 151 images per category per domain) \cite{DBLP:journals/pami/GhifaryBKZ17}. These images come from four domains: (A) AMAZON, (D) DSLR, (W) WEBCAM, and (C) CALTECH. AMAZON images were acquired in a controlled environment with studio lighting. DSLR consists of high-resolution images captured by a digital SLR camera in a home environment under natural lighting. WEBCAM images were acquired in a similar environment to DSLR but with a low-resolution webcam. CALTECH images were collected from Google Images. 
	

	We use two types of image features extracted from these datasets, \textit{i.e}.,   \textbf{SURF} and \textbf{DeCAF6}, which are publicly available.  The \textbf{SURF} \cite{gong2012geodesic} features are \textit{shallow} features extracted and quantized into an 800-bin histogram using a codebook computed with K-means on a subset of images from Amazon. The resultant histograms are further standardized by z-score. The \textbf{Deep Convolutional Activation Features (DeCAF6)} \cite{DBLP:conf/icml/DonahueJVHZTD14} are \textit{deep} features computed as in \textbf{AELM} \cite{DBLP:journals/tcyb/UzairM17}, which makes use of the VLFeat MatConvNet library with different pretrained CNN models, including in particular the Caffe implementation of \textbf{AlexNet} \cite{krizhevsky2012imagenet} trained on the ImageNet dataset. The outputs from the 6th layer are used as \textit{deep} features, leading to 4096 dimensional \textbf{DeCAF6} features. In this experiment, we denote the dataset \textbf{Amazon}, \textbf{Webcam}, \textbf{DSLR}, and \textbf{Caltech-256} as \textbf{A}, \textbf{W}, \textbf{D}, and \textbf{C}, respectively. The arrow "$\rightarrow$" is proposed to denote the direction from "source" to "target". For example, "W $\rightarrow$ D" means the Webcam image dataset is considered as the labeled \textit{source} domain, whereas the DSLR image dataset is the unlabeled \textit{target} domain.

	\textbf{USPS+MNIST} shares 10 common digit categories from two subsets, namely USPS and MNIST, which have very different data distributions (see Fig.\ref{fig:8}). We construct a first DA task \emph{USPS vs MNIST} by randomly sampling the first 1,800 images in USPS to form the source data, followed by 2,000 images in MNIST to form the target data. Then, we switch the source/target pair to get another DA task, \textit{i.e.}, \emph{MNIST vs USPS}. We uniformly rescale all images to size $16 \times 16$ and represent each one by a feature vector encoding the gray-scale pixel values. We also extract deep features from the softmax layer \cite{rozantsev2018beyond} of the LeNet \cite{lecun1998gradient} architecture, leading to a 10-dimensional feature. Thus, the source and target domain data share the same feature space. As a result, we have defined two cross-domain DA tasks, namely \emph{USPS $\rightarrow$ MNIST} and \emph{MNIST $\rightarrow$ USPS}.

	\textbf{COIL20} contains 20 objects with 1,440 images (Fig.\ref{fig:8}). The images of each object were taken by varying its pose by about 5 degrees, resulting in 72 poses per object. Each image has a resolution of 32*32 pixels and 256 gray levels per pixel. In this experiment, we partition the dataset into two subsets, namely COIL 1 and COIL 2 \cite{DBLP:journals/tip/XuFWLZ16}. COIL 1 contains all images taken within the directions in $[{0^0},{85^0}] \cup [{180^0},{265^0}]$ (quadrants 1 and 3), resulting in 720 images. COIL 2 contains all images taken in the directions
	within $[{90^0},{175^0}] \cup [{270^0},{355^0}]$ (quadrants 2 and 4) and thus the number of images is also 720. In this way, we construct two subsets with relatively different distributions. In this experiment, the COIL20 dataset with 20 classes is split into two DA tasks, \textit{i.e.},  \emph{ COIL1 $\rightarrow$ COIL2} and \emph{COIL2 $\rightarrow$ COIL1}.

	The \textbf{PIE} face database consists of 68 subjects, each under 21 various illumination conditions \cite{DBLP:journals/tip/DingF17,long2013transfer}. We adopt five pose subsets: C05, C07, C09, C27, C29, which provide a rich basis for domain adaptation, that is, we can choose one pose as the source and any remaining one as the target. Therefore, we obtain $5 \times 4=20$ different source/target combinations. Finally, we combine all five poses to form a single dataset for a large-scale transfer learning experiment. We crop all images to $32 \times 32$ and only adopt the pixel values as the input. We denote the five subsets with different face poses  as PIE1, PIE2, \textit{etc}., and generate $5 \times 4=20$ DA tasks, \textit{i.e.}, \emph{PIE1 vs PIE 2} $\dots$ \emph{PIE5 vs PIE 4}, respectively.
	
	
	\textbf{SVHN-MNIST} contains the MNIST dataset  as introduced in \textbf{USPS-MNIST} but also the Street View House Numbers (SVHN), which is a collection of house numbers collected from Google street view images (see Fig.\ref{fig:8}). SVHN is quite distinct from the dataset of handwriting digits, \textit{i.e.}, digits in MNIST. Moreover, both domains are quite large, each having at least 60k samples over 10 classes. We propose to make use of the LeNet architecture \cite{lecun1998gradient} and domain classifier as introduced in \cite{rozantsev2018beyond} to extract features for our DA tasks.

	\subsection{Baseline Methods}
	\label{subsection:Baseline Methods}
	The proposed \textbf{ARG-DA} method is compared with \textbf{thirty-three} methods from the literature, which can be differentiated according to whether they are built on the deep learning-based framework:
	

	\begin{itemize}
		\item \textbf{Shallow methods}:
		(1) 1-Nearest Neighbor Classifier(\textbf{NN}); 
		(2) Principal Component Analysis (\textbf{PCA}); 
		(3) \textbf{GFK} \cite{gong2012geodesic}; 
		(4) \textbf{TCA} \cite{pan2011domain}; 
		(5) \textbf{TSL} \cite{4967588}; 
		(6) \textbf{JDA} \cite{long2013transfer}; 
		(7) \textbf{ELM} \cite{DBLP:journals/tcyb/UzairM17}; 
		(8) \textbf{AELM} \cite{DBLP:journals/tcyb/UzairM17}; 
		(9) \textbf{SA} \cite{DBLP:conf/iccv/FernandoHST13}; 
		(10) \textbf{mSDA} \cite{DBLP:journals/corr/abs-1206-4683}; 
		(11) \textbf{TJM} \cite{DBLP:conf/cvpr/LongWDSY14}; 
		(12) \textbf{RTML} \cite{DBLP:journals/tip/DingF17}; 
		(13) \textbf{SCA} \cite{DBLP:journals/pami/GhifaryBKZ17}; 
		(14) \textbf{CDML} \cite{DBLP:conf/aaai/WangWZX14}; 
		(15) \textbf{LTSL} \cite{DBLP:journals/ijcv/ShaoKF14}; 
		(16) \textbf{LRSR} \cite{DBLP:journals/tip/XuFWLZ16}; 
		(17) \textbf{KPCA} \cite{DBLP:journals/neco/ScholkopfSM98}; 
		(18) \textbf{JGSA}  \cite{Zhang_2017_CVPR}; 
		(19) \textbf{CORAL}  \cite{sun2016return};
		(20) \textbf{RVDLR}  \cite{jhuo2012robust};
		(21) \textbf{LPJT}  \cite{li2019locality};
		(22) \textbf{DGA-DA} \cite{luo2020discriminative};
		(23) \textbf{GEF} through its different  variants, \textbf{GEF-PCA},\textbf{GEF-LDA}, \textbf{GEF-LMNN}, and \textbf{GEF-MFA} \cite{DBLP:journals/tip/ChenS0W20}.

		\item \textbf{Deep methods}: 
		(24) \textbf{AlexNet}  \cite{krizhevsky2012imagenet};
		(25) \textbf{DAH}  \cite{venkateswara2017deep}; 
		(26) \textbf{DANN}  \cite{ganin2016domain};
		(27) \textbf{ADDA} \cite{tzeng2017adversarial};
		(28) \textbf{LTRU}) \cite{sener2016learning};
		(29) \textbf{ATU} \cite{DBLP:conf/icml/SaitoUH17};
		(30) \textbf{BSWD} \cite{rozantsev2018beyond};
		(31) \textbf{DSN} \cite{bousmalis2016domain};	
		(32) \textbf{DDC} \cite{DBLP:journals/corr/TzengHZSD14};
		(33) \textbf{DAN}  \cite{long2015learning}.
	\end{itemize}
	
	A direct comparison of the proposed \textbf{ARG-DA} using shallow features against  \textbf{DL}-based DA methods could be unfair.  To address this issue, in this paper, following the previous experimental settings as reported in \textbf{DGA-DA}, \textbf{JGSA} and \textbf{BSWD}, we make use of a deep feature, namely DeCAF6, as the input feature for a fair comparison with the  \textbf{DL}-based DA methods. Whenever possible, the reported performance scores of the \textbf{thirty-three} methods from the literature are directly  collected from their original papers or previous research  \cite{tzeng2017adversarial,DBLP:journals/tcyb/UzairM17,li2019locality,DBLP:journals/pami/GhifaryBKZ17,rozantsev2018beyond,Zhang_2017_CVPR,luo2020discriminative,DBLP:journals/tip/ChenS0W20}. They are assumed to be their \emph{best} performance.


	\subsection{Experimental Setup}
	\label{subsection: Experimental setup}
	
	Given the fact that the target domain has no labeled data under the experimental setting of \textbf{UDA}, it is not possible to tune a set of optimal hyper-parameters. In this paper, following the setting of previous research \cite{luo2020discriminative,long2013transfer,DBLP:journals/tip/XuFWLZ16}, we also evaluate the proposed \textbf{ARG-DA} by empirically searching the parameter space for the \emph{optimal} settings. Specifically, the proposed \textbf{ARG-DA} method has three hyper-parameters, \textit{i.e.}, the subspace dimension $k$, and regularization parameters $\lambda $ and $\alpha $. In our experiments, we set $k = 200$ and 1) $\lambda= 0.1$, and $\alpha  = 0.9$ for \textbf{USPS}, \textbf{MNIST}, \textbf{COIL20}, and \textbf{PIE}, 2) $\lambda  = 1$, $\alpha  = 0.9$ for \textbf{Office+Caltech}, and \textbf{SVHN-MNIST}.
	
	In our experiment, {\emph{accuracy}} on the test dataset as defined by Eq.(\ref{eq:accuracy}) is used as the performance measurement. It is widely applied in a series of previous research work, \textit{e.g.}, \cite{long2015learning,DBLP:journals/corr/LuoWHC17,long2013transfer,DBLP:journals/tip/XuFWLZ16}, \textit{etc}.
	
	\begin{equation}\label{eq:accuracy}
		\begin{array}{c}
			Accuracy = \frac{{\left| {x:x \in {D_T} \wedge \hat y(x) = y(x)} \right|}}{{\left| {x:x \in {D_T}} \right|}}
		\end{array}
	\end{equation}
	where ${\cal{D_T}}$ is the target domain treated as test data, ${\hat{y}(x)}$ is the predicted label, and ${y(x)}$ is the ground truth label for the test data  $x$.

	As discussed in Sect.\ref{subsection:Formulation}, the core model of the proposed \textbf{ARG-DA} method is originated from \textbf{DGA-DA} but combines two optimization strategies, namely the Attention Regularized Laplace Graph (\textbf{ARG}) term as defined in Eq.(\ref{eq:12}), and the Joint Manifold Alignment Strategy ('\textbf{FEEL}') as introduced in Sect.\ref{ARG-based Manifold Alignment}. To gain insight into the proposed \textbf{ARG-DA} and the rationale \textit{w.r.t} the \textbf{ARG} term and the \textbf{FEEL} strategy, we derive from Eq.(\ref{eq:18}) one intermediate partial model, namely \textbf{DGA+A}, to highlight the contribution of each optimization strategy to the final proposed \textbf{ARG-DA}:


	\begin{itemize}
		\item \textbf{DGA+A}: in this setting,  we remove ${{\mathbf{AR}}{{\mathbf{G}}^O}}$ from Eq.(\ref{eq:18}) and derive the \textbf{DGA+A} model, which only makes use of the \textbf{ARG} mechanism as defined in Eq.(\ref{eq:12}) but without the '\textbf{FEEL}' strategy as defined in Sect.\ref{ARG-based Manifold Alignment}. This setting makes it possible to understand the contribution of the proposed \textbf{ARG} mechanism when the different sub-domain adaptation is properly regularized to remedy one of the raised issues, namely '\textbf{Lack of Attention}', by comparing it with its baseline \textbf{DA} model \textbf{DGA-DA}. Specifically, the mathematical formulations of both \textbf{DGA+A} and \textbf{DGA-DA} are depicted in Fig.\ref{fig:18} for better clarity.

		\item \textbf{ARG-DA}: now our proposed \textbf{ARG-DA} improves \textbf{DGA+A} by further implementing the  '\textbf{FEEL}' strategy to additionally align the original feature manifold for comprehensive manifold unification. With \textbf{ARG+A} as baseline, it is now possible to highlight the contribution of the '\textbf{FEEL}' strategy when all the manifold structures are further aligned.

	\end{itemize}

	\subsection{Experimental Results and Discussion}
	\label{subsection: Experimental Results and Discussion}

	\subsubsection{\textbf{Experiments on the CMU PIE Data Set}}
	\label{subsubsection:Experiments on the CMU PIE dataset}
	
\begin{figure*}[h!]
		\centering
		\includegraphics[width=1\linewidth]{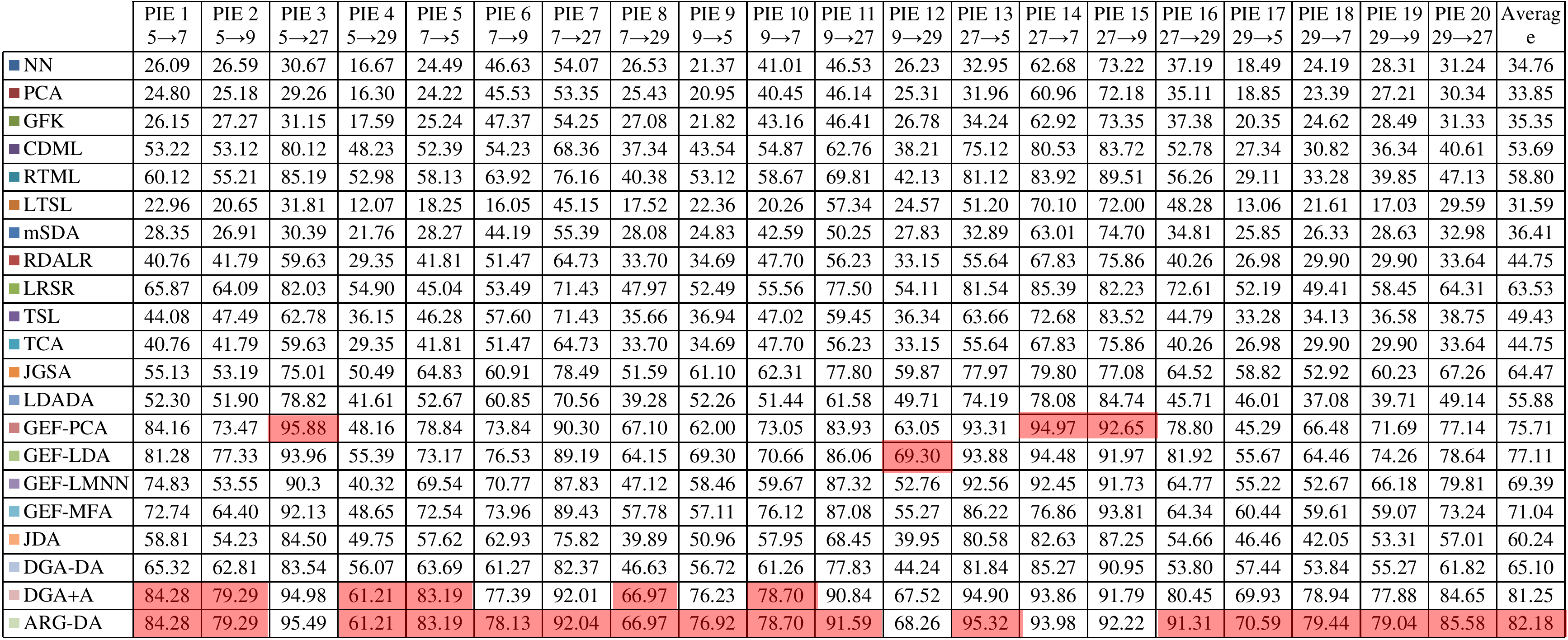}
		\caption {Accuracy${\rm{\% }}$ on the PIE Images Dataset.} 
		\label{fig:9}
		\vspace{-1em}
	\end{figure*}

	The \textbf{CMU PIE} database is a large face dataset containing 68 people with different pose, illumination, and expression variations. Fig.\ref{fig:9} synthesizes the experimental results for \textbf{DA} using this dataset, where the best results are highlighted in red. The following observations can be drawn from Fig.\ref{fig:9}:
	
	
	\begin{itemize}
		
		\item by implementing the Attention Regularized Laplace Graph Enforced Adaptation model, \textbf{DGA+A} enhances discriminativeness across different  sub-domains, thereby improving its baseline \textbf{DGA-DA} by 16 points, and achieves $81.25\%$ accuracy. By making use of the \textbf{FEEL} strategy, \textbf{ARG-DA} additionally improves \textbf{DGA+A} by roughly 1 point and highlights the interest of unifying the alignment of different feature/label manifolds.

		\item Interestingly, the second best performing methods are \textbf{GEF-PCA} and \textbf{GEF-LDA}, respectively. They are both variants of \textbf{GEF}  \cite{DBLP:journals/tip/ChenS0W20}. \textbf{GEF-PCA}, formalized as Eq.(\ref{eq:26}), improves the baseline \textbf{JDA} by 15 points by incorporating a specifically designed revision graph ${\Delta {\mathbf{G}}}$ in order to increase within-class compactness.

		\begin{equation}\label{eq:26}
			\resizebox{0.75\hsize}{!}{%
				$\mathop {\min }\limits_{{{\mathbf{A}}^T}{\mathbf{XH}}{{\mathbf{X}}^T}{\mathbf{A}} = {\mathbf{I}}} (\sum\limits_{c = 0}^C {tr({{\mathbf{A}}^T}{\mathbf{X}}({\mathbf{M}} + \Delta {\mathbf{G}}){{\mathbf{X}}^T}{\mathbf{A}})}  + \lambda \left\| {\mathbf{A}} \right\|_F^2)$}
		\end{equation}
		

		\item On the other hand, \textbf{GEF-LDA} draws inspiration from Linear Discriminant Analysis (\textbf{LDA}) \cite{DBLP:journals/pami/MartinezK01}, and replaces the constraint ${{{\mathbf{A}}^T}{\mathbf{XH}}{{\mathbf{X}}^T}{\mathbf{A}} = {\mathbf{I}}}$ in \textbf{GEF-PCA} by ${{{\mathbf{A}}^T}{\mathbf{XP}}{{\mathbf{X}}^T}{\mathbf{A}} = {\mathbf{I}}}$. Also, it further leverages the supervised knowledge embedded in the graph matrix $\mathbf{P}$ in order to minimize within-class scatter and maximize inter-class scatter, thereby further improving \textbf{GEF-PCA} by 1 point. 
		
		\item  The proposed \textbf{ARG-DA} is different from \textbf{GEF-LDA}. \textbf{ARG-DA} explicitly enhances inter-class discriminativeness through the designed \textit{repulsive force} term as in Eq.(\ref{eq:CDDAnew}), while the effectiveness of within-class compactness as highlighted in \textbf{GEF-PCA} is potentially enforced by dynamically regressing the common feature representation to the discriminative label space through the proposed \textbf{ARG} term (Sect.\ref{Attention Regularized Laplace Graph}) regularized \textbf{DGA+A}. Finally, \textbf{ARG-DA} improves \textbf{DGA+A} by additionally enabling the smooth manifold alignment across different feature spaces and thus achieves the best accuracy of $81.18\%$.

	\end{itemize}



	\subsubsection{\textbf{Experiments on the USPS+MNIST Dataset}}
	\label{subsubsection: experiments on the UPS+MNIST Datasets}
	
	\begin{figure*}[h!]
		\centering
		\includegraphics[width=1\linewidth]{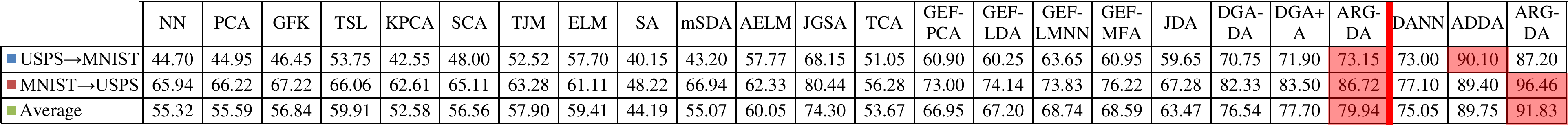}
		
		\caption {Accuracy${\rm{\% }}$ on the USPS+MNIST Image Dataset.} 
		\label{fig:10}
		\vspace{-2em}
	\end{figure*} 
	
	The \textbf{USPS+MNIST} dataset displays different writing styles between the source and target domains. In Fig.\ref{fig:10}, the left-hand columns of the red vertical bar report the experimental results using shallow features, whereas the right-hand columns of the red bar display the results of the methods using deep features.

	\begin{itemize}

		\item Using shallow features, \textbf{DGA-DA} outperforms both \textbf{GEF-PCA} and \textbf{GEF-LDA} by a large margin on the \textbf{USPS+MNIST} dataset. This is in contrast with the previous results in Sect.\ref{subsubsection:Experiments on the CMU PIE dataset} where \textbf{GEF-PCA} and \textbf{GEF-LDA} achieve higher accuracy than \textbf{DGA-DA}. We suspect that the main reason is due to the random writing styles of the \textbf{USPS+MNIST} dataset, which generates a more sophisticated data manifold structure than the one by predefined pose variations of the \textbf{PIE} dataset. As a result, by explicitly capturing the underlying geometric structures through the Laplace graph and leveraging manifold learning  (${{\mathbf{Y}}^T}{\mathbf{LY}}$), as well as the label smoothness consistency (Eq.(\ref{eq:16})) for robust label propagation, \textbf{DGA-DA} seems better armed to solve the complex manifold structure induced by \textbf{DA} tasks.

		\item By re-weighting the corresponding distances across different sub-domains for intra-class and inter-class aware  \textbf{DA}, \textbf{DGA+A} improves \textbf{DGA-DA} by 1 point and achieves $77.70\%$ accuracy. By taking advantage of the '\textbf{FELL}' strategy for comprehensive manifold learning, \textbf{ARG-DA} further improves \textbf{DGA+A} and achieves $79.94\%$ accuracy.


		\item In the right-hand part of the red vertical bar in Fig.\ref{fig:10}, we also compare our approach with deep network-based methods, \textit{i.e.}, \textbf{ADDA} and \textbf{DANN}, which search a common latent subspace by shrinking the domain divergence using the adversarial learning strategy, and improve the performance of traditional \textbf{DA} methods by a large margin. Using deep features \cite{rozantsev2018beyond}, \textbf{ARG-DA} demonstrates its effectiveness and outperforms deep learning-based \textbf{DA} methods, \textit{e.g.}, \textbf{ADDA} and \textbf{DANN}, by a margin of 16 and 2 points, respectively.

	\end{itemize}

	\subsubsection{\textbf{Experiments on the Office+Caltech-256 Datasets}}
	\label{subsubsection:Experiments on the Office+Caltech-256 Data Sets}
	
	\begin{figure}[h!]
		\centering
		\includegraphics[width=1\linewidth]{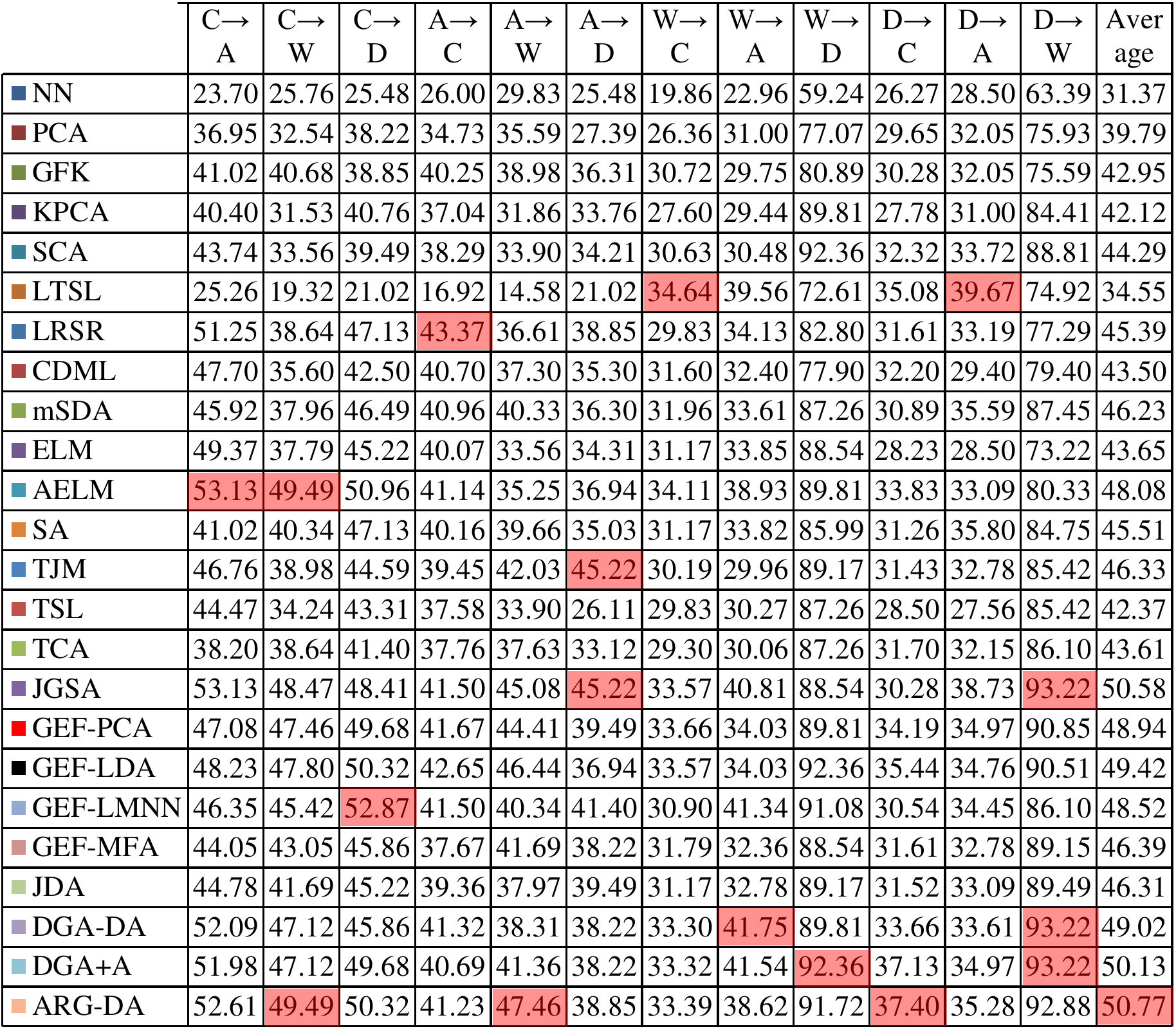}
		
		\caption {Accuracy${\rm{\% }}$ on the Office+Caltech Images with SURF-BoW Features.} 
		\label{fig:11}

	\end{figure} 
	
	\begin{figure}[h!]
		\centering
		\includegraphics[width=1\linewidth]{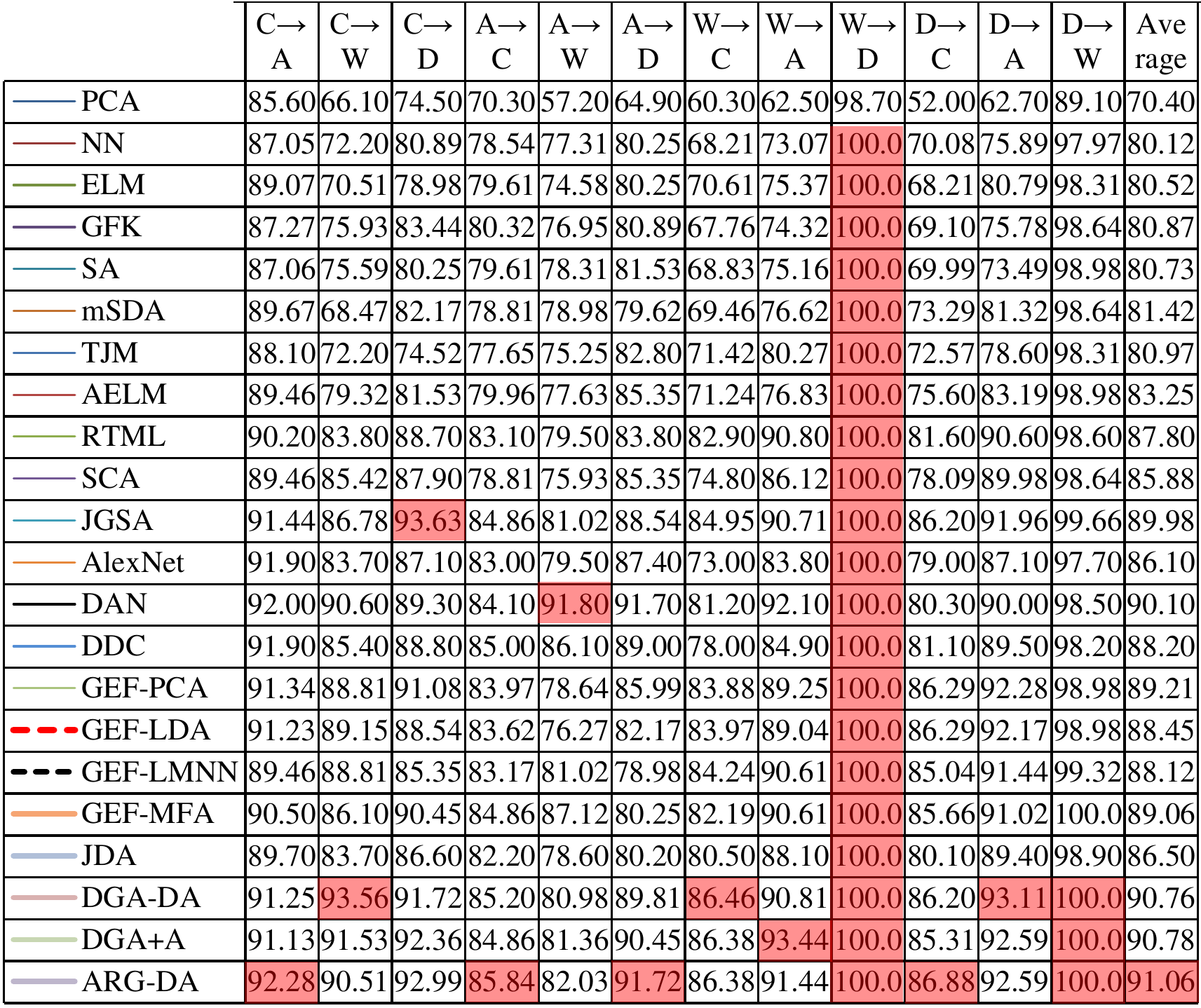}
		
		\caption {Accuracy${\rm{\% }}$ on the Office+Caltech Images with DeCAF6 Features.} 
		\label{fig:12}
		\vspace{-1em}
	\end{figure}

	Fig.\ref{fig:11} and Fig.\ref{fig:12} synthesize the experimental results in comparison with the state-of-the-art when classic shallow features (\textit{i.e.}, SURF features) and deep features (\textit{i.e.}, DeCAF6 features) are used, respectively. From Fig.\ref{fig:11} and Fig.\ref{fig:12}, the following observations can be drawn: 
	
	\begin{itemize}
		\item In both Fig.\ref{fig:11} and Fig.\ref{fig:12}, \textbf{DGA+A} outperforms \textbf{DGA-DA}, which suggests the effectiveness of \textbf{ARG} enforced attention aware \textbf{DA} \textit{w.r.t.} different feature representations. Meanwhile, it can be observed that \textbf{ARG-DA} improves \textbf{DGA+A} in both Fig.\ref{fig:11} and Fig.\ref{fig:12}, thereby further demonstrating the contribution of the '\textbf{FEEL}' strategy to comprehensive manifold learning.


		\item In Fig.\ref{fig:12} using deep features, both \textbf{DGA+A} and \textbf{ARG-DA} improve their performance as reported in Fig.\ref{fig:11} by a very large margin. These results demonstrate that deep features can be seamlessly embedded into the proposed \textbf{DA} models for improved performance. 

	\end{itemize}

	\subsubsection{\textbf{Large-Scale Experiments on the SVHN-MNIST Dataset}}
	\label{subsubsection:Experiments on the SVHN-MNIST dataset}
	
\begin{figure}[h!]
		\centering
		\includegraphics[width=1\linewidth]{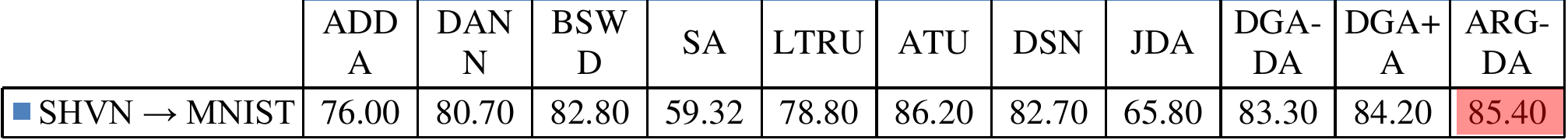}
		
		\caption {Accuracy${\rm{\% }}$ on the SVHN-MNIST Images Dataset.} 
		\label{fig:13}
	\end{figure}

	Different from the other datasets, \textbf{SVHN-MNIST} is composed of 50k and 20k real-world images from two very different domains, respectively, thereby generating a large-scale \textbf{DA} benchmark. Following the same experimental setting of previous research \cite{luo2020discriminative},  Fig.\ref{fig:13} reports the final experimental results. As shown in Fig.\ref{fig:13}, thanks to the effectiveness of the \textbf{ARG} term enforced manifold embeddings,  \textbf{DGA+A} improves \textbf{DGA-DA} by 1 point. Furthermore, by additionally embracing the \textbf{FEEL} strategy ensured comprehensive manifold learning techniques, \textbf{ARG-DA} further improves \textbf{DGA+AA} by an accuracy of 1 point. In Fig.\ref{fig:13}, \textbf{ARG-DA} achieves the best performance, even in comparison with a series of state-of-the-art deep learning-based \textbf{DA} methods,  thereby further demonstrating its effectiveness.


	\subsubsection{\textbf{Experiments on the COIL 20 Dataset}} 
	\label{subsubsection: results on the COIL dataset}
	
	The \textbf{COIL} dataset (see fig.\ref{fig:8}) features the challenge of pose variations between the source and target domains. Even though  \textbf{DGA-DA}  already achieves $100.00\%$ accuracy on the \textbf{COIL} dataset, we still benchmark our proposed \textbf{ARG-DA} in order to see how the proposed \textbf{ARG} term and \textbf{FEEL} strategy impact the best baseline performance. Fortunately, both \textbf{DGA+A} and \textbf{ARG-DA} achieve $100.00\%$  accuracy and verify the robustness of the proposed \textbf{ARG} term and \textbf{FEEL} strategy.


	\subsection{\textbf{Empirical Analysis}}
	\label{Empirical Analysis}

	Although the proposed \textbf{ARG-DA} achieves state-of-the-art performance over 37 DA tasks through 7 datasets, an interesting question is how fast the proposed method converges (sect.\ref{Convergence analysis}), as well as its sensitivity \textit{w.r.t.} its hyper-parameters (Sect.\ref{Variants of Parameters}). Additionally, in Sect.\ref{comprehensive manifold learning}, we also aim to explore the robustness of the proposed \textbf{FEEL} strategy (Sect.\ref{ARG-based Manifold Alignment}) without the collaboration of the \textbf{ARG} term.

	\subsubsection{\textbf{Sensitivity of the proposed ARG-DA \textit{w.r.t.} hyper-parameters}}
	\label{Variants of Parameters} 
	
	Three hyper-parameters, namely $k$, $\lambda$ and $\alpha$, are used in the proposed \textit{DA} method.

	\begin{itemize}
		\item $k$ denotes the dimension of the searched shared latent feature subspace, which determines the structure of low-dimension embedding. Obviously, the larger $k$, the more the shared subspace can afford complex data distributions, but at the cost of increased computation complexity.
		
		\item $\lambda$ defined in Eq.(\ref{eq:24}) aims to regularize the projection matrix $A$ to avoid over-fitting the chosen shared feature subspace \textit{w.r.t.} both the source and the target domain. Increasing $\lambda$ reduces the risk of model over-fitting but shrinks the diversity of the functional learning representations.
		
		\item $\alpha {\rm{ = }}\frac{1}{{1 + \mu }}$ as defined in Eq.(\ref{eq:24}) is a trade-off parameter, which balances \textbf{LSmC} term (Eq.(\ref{eq:16})) and label space embedding (Eq.(\ref{eq:15})). Thus, increasing $\alpha$ improves the effectiveness of label space embedding, while reducing the contribution of label smoothness regularization.
		
	\end{itemize}

\begin{figure}[h!]
		\centering
		\includegraphics[width=1\linewidth]{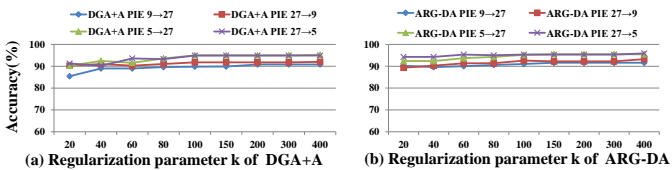}
		
		\caption {Sensitivity analysis of the proposed methods using the PIE dataset:  (a) accuracy \textit{w.r.t.} subspace dimension $k$ of \textbf{DGA+A}; (b) accuracy \textit{w.r.t.} subspace dimension $k$ of \textbf{ARG-DA}.} 
		\label{fig:15}
	\end{figure}

	\textbf{Sensitivity of $k$:} in Fig.\ref{fig:15}, we plot the classification accuracies of the proposed \textbf{DA} method \textit{w.r.t} different values of $k$ on the \textbf{PIE} dataset. As shown in Fig.\ref{fig:15}, the subspace dimensionality $k$ varies with $k \in \{20,40,60,80,100,150,200,300,400\}$. The proposed two \textbf{DA} variants, namely, \textbf{DGA+A} and \textbf{ARG-DA}, remain stable \textit{w.r.t.} a wide range of $k \in \{ 60 \le k  \le 400\} $. Finally, in this paper,  we set $k =200$ to balance efficiency and accuracy.

\begin{figure}[h!]
		\centering
		\includegraphics[width=1\linewidth]{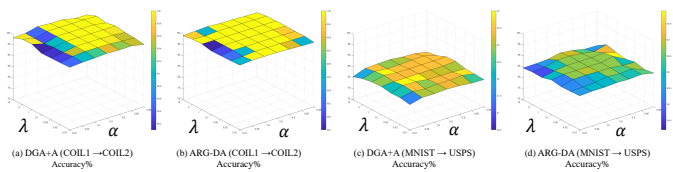}
		
		\caption {The classification accuracies of the proposed \textbf{ARG-DA} and \textbf{DGA+A} methods vs.the parameters $\alpha $ and $\lambda $ on the selected three cross-domain datasets.} 
		\label{fig:16}

	\end{figure}

	\textbf{Sensitivity of $\lambda$ and $\alpha$:} we also study the sensitivity of the proposed \textbf{DGA+A} and \textbf{ARG-DA} methods with a wide range of hyper-parameter values, \textit{i.e.}, $\alpha  = (0.01,0.05,0.1,0.5,0.9,0.95,0.99)$ and $\lambda  = (0.001,0.002,0.005,0.1,1,5,10)$. We plot in Fig.\ref{fig:16} the results on \emph{COIL1} $\rightarrow$ \emph{COIL2}  $ and $  \emph{MNIST} $\rightarrow$ \emph{USPS} datasets on both methods with $k$ held fixed at $200$. As shown in Fig.\ref{fig:16}, the proposed \textbf{DGA+A} and \textbf{ARG-DA} display their stability as the resultant classification  accuracy remains roughly the same despite a wide range of $\lambda $ and $\alpha $ values, thereby suggesting that the proposed methods can easily search the proper candidate for these hyper-parameters.

		\vspace{1em}
	\subsubsection{\textbf{Convergence analysis}}
	\label{Convergence analysis}
	
	\begin{figure}[h!]
		\centering
		\includegraphics[width=1\linewidth]{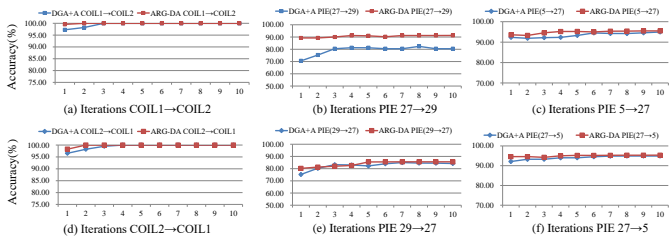}
		
		\caption {Convergence analysis using 6 cross-domain image classification tasks on COIL and PIE datasets. (accuracy w.r.t $\#$iterations)} 
		\label{fig:17}
	\end{figure}


   Another interesting question is to see when the proposed methods  achieve their best performance \textit{w.r.t.} the number of iterations $T$. For this purpose, in Fig.\ref{fig:17}, we carry out a  convergence analysis of the proposed \textbf{DGA+A} and \textbf{ARG-DA} models, using the pixel value features on both the \textbf{PIE} and \textbf{COIL} datasets.  Subsequently, Fig.\ref{fig:17} reports 6 cross DA experiments ( \emph{COIL1} $\rightarrow$ \emph{2}, \emph{COIL2} $\rightarrow$ \emph{COIL1} ...  \emph{PIE-27} $\rightarrow$ \emph{PIE-5}  ) with the number of iterations $T = (1,2,3,4,5,6,7,8,9,10)$. As can be seen in Fig.\ref{fig:17}, both \textbf{DGA+A} and \textbf{ARG-DA} converge within 3$ \sim $5 iterations when performing optimization over the two datasets. On the other hand, \textbf{ARG-DA} requires slightly more computational time per iteration than \textbf{DGA+A}, due to the computation of the additionally designed ${{\mathbf{AR}}{{\mathbf{G}}^O}}$, which discriminatively aligns different feature spaces. However, thanks to the \textbf{FEEL} strategy, the convergence curve displayed by \textbf{ARG-DA} is flatter in  comparison with \textbf{DGA+A}.

	\begin{figure}[h!]
		\centering
		\includegraphics[width=1\linewidth]{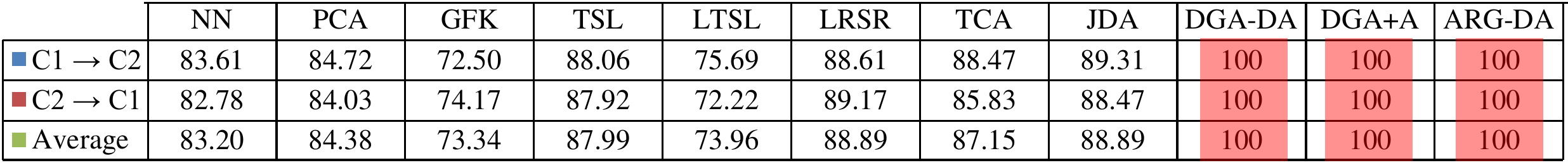}
		
		\caption {Accuracy${\rm{\% }}$ on the COIL Images Dataset.} 
		\label{fig:14}

	\end{figure}

	\subsubsection{\textbf{Robustness of the \textbf{FEEL}-based comprehensive manifold learning}}
	\label{comprehensive manifold learning}
	
	The \textbf{FEEL} strategy proposed in the \textbf{ARG-DA} model consists in aligning the manifold structures (\textbf{MS}) across the following groups of feature/label spaces:

	\begin{itemize}
		\item (\textbf{G1}). manifold structure across the original feature space (${\mathcal{X}}$) and the optimized common feature space (${\mathbf{A}}*{\mathcal{X}}$);
		
		\item (\textbf{G2}). manifold structure across the optimized feature space (${\mathbf{A}}*{\mathcal{X}}$) and the label space (${\mathcal{Y}}$);
	\end{itemize}	
	
	Alignment of \textbf{G1} or \textbf{G2}  in \textbf{DA} has already shown their effectiveness in a series of current research work, thereby encouraging hybridization of the previous two approaches by aligning jointly \textbf{G1} and \textbf{G2} within a unified optimization framework, as is the case in the proposed \textbf{FEEL} strategy. Even though the contribution of the \textbf{FEEL} strategy has already been highlighted by contrasting \textbf{ARG-DA} with its baseline model, \textbf{DGA+A}, over various datasets, we are still curious about the effectiveness of the \textbf{FEEL} strategy when it is disentangled from \textbf{ARG} regularization as in our previously proposed \textbf{DGA-DA} model. For this purpose, we derive a novel partial model, namely, \textbf{DGA+F}, which is  formulated in Eq.(\ref{eq:27})


	\begin{equation}\label{eq:27}
		\resizebox{0.6\hsize}{!}{%
			$\min \left( {\begin{array}{*{20}{l}}
				{\sum\limits_{c = 0}^C {tr({{\mathbf{A}}^T}{\mathbf{K}}({{\mathbf{M}}^*} + {\mathbf{L'}}){{\mathbf{K}}^T}{\mathbf{A}})}  + \lambda \left\| {\mathbf{A}} \right\|_F^2} \\ 
				{ + \mu (\sum\limits_{j = 1}^C {\sum\limits_{i = 1}^{{n_s} + {n_t}} {\left\| {{\mathbf{Y}}_{ij}^{(F)} - {\mathbf{Y}}_{ij}^{(0)}} \right\|} } ) + {{\mathbf{Y}}^T}{\mathbf{LY}}} 
				\end{array}} \right){\text{ }}$}
	\end{equation}
	
\begin{figure}[h!]
		\centering
		\includegraphics[width=1\linewidth]{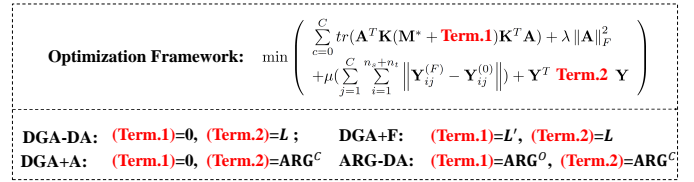}
		
		\caption {Detailed discussion of the differences across the \textbf{DGA-DA}, \textbf{DGA+F}, \textbf{DGA+A}, and \textbf{ARG-DA} algorithms.} 
		\label{fig:18}
	\end{figure}

	Where ${{\mathbf{L'}}}$ and ${\mathbf{L}}$ intend to align the manifold structures of \textbf{G1} and \textbf{G2}, respectively. To gain insight into the proposed \textbf{FEEL} strategy, we draw the general optimization framework in Fig.\ref{fig:18}, where four different experimental settings, \textit{e.g.},  \textbf{DGA-DA}, \textbf{DGA+F}, \textbf{DGA+A}, and \textbf{ARG-DA}, are clearly distinguished by 
	varying  \textbf{Term.1} and \textbf{Term.2} of the optimization framework.


\begin{figure}[h!]
		\centering
		\includegraphics[width=1\linewidth]{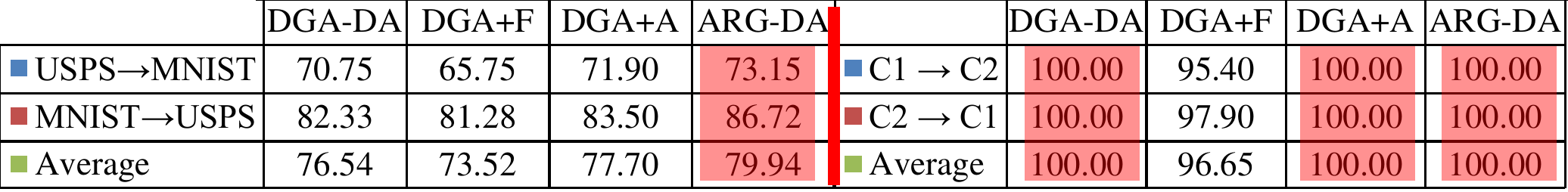}
		
		\caption {Accuracy${\rm{\% }}$ on the USPS+MNIST Images Dataset using different experimental settings.} 
		\label{fig:19}
	\end{figure}

	Fig.\ref{fig:19} shows the performance of the four experimental settings using the pixel value features on both the \textbf{USPS+MNIST} and \textbf{COIL} datasets.  As shown in Fig.\ref{fig:19}, the following can be observed:

	\begin{itemize}
		\item Mathematically, \textbf{DGA+F} is different from \textbf{DGA-DA} by  updating \textbf{Term.1} as ${\mathbf{L'}}$, which intends to additionally align the manifold structure across \textbf{G1}. However, \textbf{DGA+F} underperforms in comparison with its baseline, \textbf{DGA-DA}, both on the \textbf{USPS+MNIST} and \textbf{COIL} datasets.

		
		\item Under the condition of using the \textbf{ARG} term-based regularization, \textbf{ARG-DA} improves \textbf{DGA+A} by incorporating the \textbf{FEEL} strategy, which can be seen on both the \textbf{USPS+MNIST} and \textbf{COIL} datasets, in line with the results analyzed in Sect.\ref{subsubsection:Experiments on the CMU PIE dataset} through Sect.\ref{subsubsection:Experiments on the SVHN-MNIST dataset}.

		
	\end{itemize}	
	
	To conclude, the \textbf{FEEL} strategy cannot be simply incorporated into the \textbf{DGA-DA} model. Due to the manifold structure divergence caused by domain shift ,  brutally aligning the manifold structure across \textbf{G1} hurts the final \textbf{DA} performance. In this paper, by collaborating with the \textbf{ARG} term, the \textbf{FEEL} strategy is seamlessly embedded into the \textbf{DGA+A} model to dynamically align the manifold structure across different feature/label spaces and eventually achieve comprehensive manifold learning.


		\vspace{-2em}

	\section{Conclusion}
	\label{Conclusion}
    In this paper, we proposed a novel unsupervised \textbf{DA} method, namely \textbf{A}ttention \textbf{R}egularized Laplace \textbf{G}raph-based \textbf{D}omain \textbf{A}daptation (\textbf{ARG-DA}), which provides new perspectives, \textit{i.e.}, class aware attention and manifold alignment, to further explore Laplace graph-based \textbf{DA} methods. By weighting the importance of different sub-domain adaptations, we designed the \textbf{A}ttention \textbf{R}egularized Laplace \textbf{G}raph to enforce the attention aware \textbf{DA} model \textbf{DGA+A}. Moreover, by unifying knowledge across different feature/label spaces, we improved \textbf{DGA+A} by incorporating the \textbf{FEEL} strategy for comprehensive manifold learning, thereby providing our final \textbf{ARG-DA}. 
	Comprehensive experiments using the standard benchmark  in  \textbf{DA}  showed  the  effectiveness  of the proposed method, which consistently outperforms state-of-the-art \textbf{DA} methods. Future work includes embedding of the proposed \textbf{ARG-DA} into the paradigm of deep learning in order to improve the real computer vision applications, \textit{e.g.}, detection, segmentation, and tracking, \textit{etc}.

\ifCLASSOPTIONcaptionsoff
  \newpage
\fi

\small\bibliographystyle{plain}
\small\bibliography{cdda}

	\vspace{-5em}

\begin{IEEEbiography}   [{\includegraphics[width=1in,height=1.25in,clip,keepaspectratio]{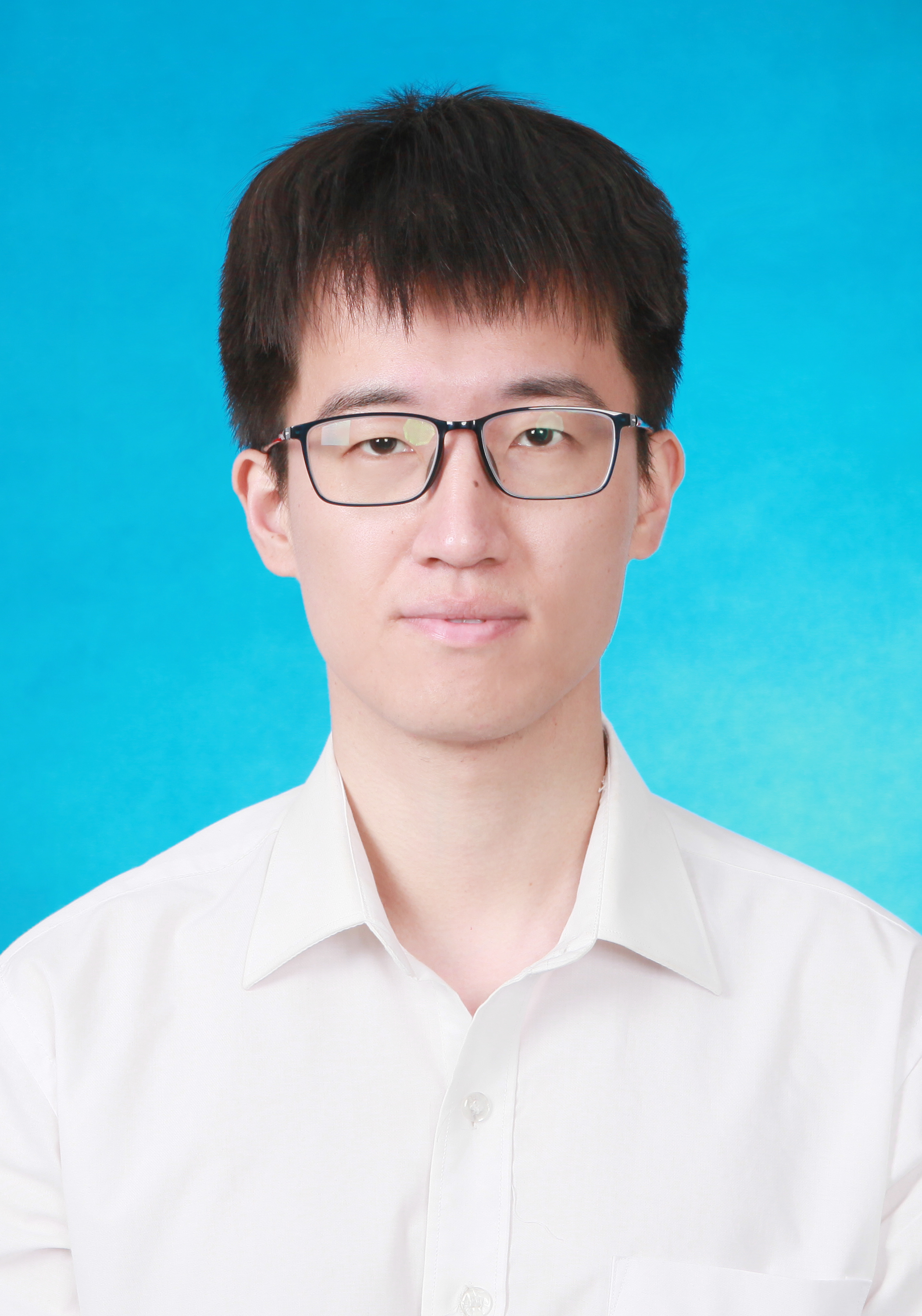}}]{Lingkun Luo}
	was awarded his Ph.D degree at the Shanghai Jiao Tong University. He served as research assistant and postdoc in the Ecole Centrale de Lyon, Department of Mathematics and Computer Science, and was a member of the LIRIS laboratory. Today, he is a research fellow of the Shanghai Jiao Tong University. He has authored more
	than 20 research papers. His research interests include machine learning, pattern recognition, and computer vision.
\end{IEEEbiography}

	\vspace{-1em}
\begin{IEEEbiography}[{\includegraphics[width=1in,height=1.25in,clip,keepaspectratio]{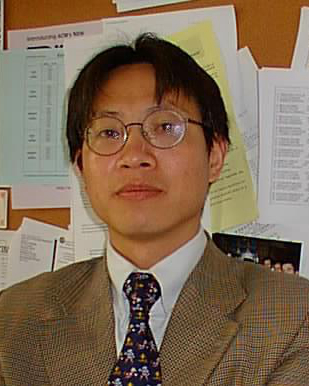}}]{Liming Chen} was awarded a joint B.Sc. degree in mathematics and computer science from the University of Nantes, Nantes, France in 1984, as well as M.Sc. and Ph.D. degrees in computer science from the University of Paris 6, Paris, France, in 1986 and 1989, respectively.
	
	He first served as an Associate Professor with the Universit\'{e} de Technologie de Compi\`{e}gne, before joining the \'Ecole Centrale de Lyon, \'Ecully, France, as a Professor in 1998, where he leads an advanced research team on Computer Vision, Machine Learning, and Multimedia. From 2001 to 2003, he also served as Chief Scientific Officer in a Paris-based company, Avivias, specializing in media asset management. In 2005, he served as Scientific Multimedia Expert for France Telecom R\&D China, Beijing, China. He was the Head of the Department of Mathematics and Computer Science in the \'Ecole Centrale de Lyon from 2007 through 2016. His current research interests include computer vision, machine learning, image and multimedia, with a particular focus on robot vision and learning since 2016. Liming has authored over 300 publications and has successfully supervised more than 40 Ph.D students. He has been a grant holder for a number of research grants from the EU FP program, French research funding bodies, and local government departments. Liming has so far guest-edited 5 journal special issues. He is an associate editor for the Eurasip Journal on Image and Video Processing and a senior IEEE member.
\end{IEEEbiography}

\begin{IEEEbiography}   [{\includegraphics[width=1in,height=1.25in,clip,keepaspectratio]{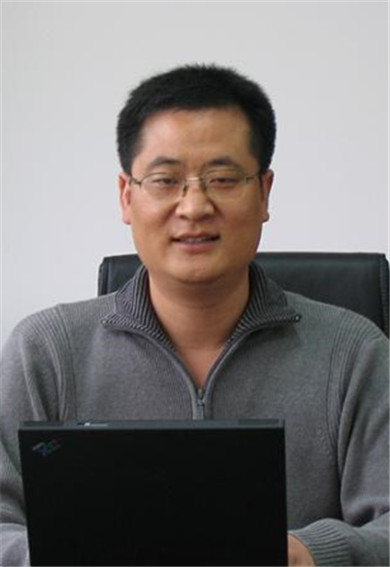}}]{Shiqiang Hu}
	
	was awarded his Ph.D degree at the Beijing Institute of Technology. He has authored over 150 publications and has successfully supervised more than 15 Ph.D students. Today, he is a full professor at the Shanghai Jiao Tong University. His research interests include data fusion technology, image understanding, and nonlinear filter.
\end{IEEEbiography}

\end{CJK*}
\end{document}